\RequirePackage{fix-cm}

\documentclass[10pt,twocolumn,twoside,journal]{article}
\usepackage[papersize={216mm,285mm},textheight=234mm,left=21mm, right=21mm, top=27mm, columnsep=5.5mm,footskip=9mm]{geometry}
\usepackage[small,bf]{caption}

\usepackage{subcaption}
\usepackage{amsmath}
\usepackage{amssymb}
\usepackage{makecell}
\usepackage{placeins}
\usepackage{bm}
\usepackage{authblk}
\usepackage{graphicx}
\usepackage{url}
\usepackage{multirow}
\usepackage{amsmath,amsthm}
\usepackage{amssymb,amsfonts}
\usepackage{booktabs}
\usepackage{color}
\usepackage{booktabs}
\usepackage{float}
\usepackage{fancyhdr}
\usepackage{xcolor,stfloats}
\usepackage{comment}
\usepackage{cuted}  
\usepackage{epstopdf}
\usepackage{multicol}
\usepackage{algorithmic}
\usepackage{algorithm}
\def\ManuscriptInfo{Manuscript received: ; accepted: }

\title{Imposing Temporal Consistency on Deep Monocular Body Shape and Pose Estimation}
\author[1,2]{A. Zimmer}
\author[1]{A. Hilsmann}
\author[1]{W. Morgenstern}
\author[1,3]{P. Eisert}
\affil[1]{Fraunhofer HHI}
\affil[2]{TU Berlin}
\affil[3] {HU Berlin}
\date{}

\begin{document}
\maketitle

\begin{abstract}{Accurate and temporally consistent modeling of human bodies
is essential for a wide range of applications, including character animation, understanding human social behavior and AR/VR interfaces. 
Capturing human motion accurately from a monocular image sequence is still challenging and 
the modeling quality is strongly influenced by the temporal consistency of the captured body motion.
Our work presents an elegant solution for the integration of temporal constraints in the fitting process. This does not only increase temporal consistency but also robustness during the optimization.
In detail, we derive parameters of a sequence of body models, representing shape and motion of a person, including jaw poses, facial expressions, and finger poses. We optimize these parameters over the complete image sequence, fitting one consistent body shape while imposing temporal consistency on the body motion, assuming linear body joint trajectories over a short time. Our approach enables the derivation of realistic 3D body models from image sequences, including facial expression and articulated hands.
In extensive experiments, we show that our approach results in accurately estimated body shape and motion, also for challenging movements and poses. Further, we apply it to the special application of sign language analysis, where accurate and temporal consistent motion modelling is essential, and show that the approach is well-suited for this kind of application.
}
\end{abstract}

\section{Introduction}\label{sec:intro}

Estimation of human 3D body motion and shape from monocular image sequences and videos is a challenging task in the field of computer vision and graphics. 
There are many approaches addressing this problem, but most of them estimate motion and shape on a frame by frame basis without addressing temporal consistency, 
and thus suffer from artifacts and inaccurate results. 
There is e.g. SMPL-X \cite{SMPLify-X}, which uses a parametric human shape model that is fitted to the individual frames, without explicitly accounting for temporal consistency, and which is prone to inconsistent shape and pose, leading to wiggling and jittering in the temporal sequence. 
These methods ignore the information of temporal vicinity present in a given image sequence or video. However, this information should be used to improve body pose and shape estimation. 

The approach presented in this paper accurately estimates human body motion and shape from monocular image sequences, and adds temporal consistency constraints for more stable solutions. 
It finds a consistent body shape for the entire image sequence and imposes temporal consistency on the body pose estimates from image to image. 
Our approach includes the estimation of facial expression and finger poses, which makes it suitable for highly detailed body motion estimation, such as sign language capture.

Possible applications of body shape and pose estimation lie in body motion tracking, character animation, understanding human social behavior and AR/VR interfaces in favor of the entertainment industry, robotics, sports analytics, medical applications and many more.
Another possible application of our approach is the generation of realistic sign language using avatars. Currently, not all media provide additional sign language. Therefore, human sign language interpreters are needed.
Automatically generated avatars, which are able to realistically and human-like emulate detailed sign language, would enable barrier-free access to all acoustically supported media. 
A current state of the art approach for generating sign language motion is the employment of a deep neural network to learn from large amounts of sign language training data.
With the results of this paper, such training data can be acquired 
in large quantities by accurately estimating sign language body motion from videos or image sequences. 

We evaluated our body pose and shape estimation approach on image sequences and videos with simple and complex body poses and motion, including sign language image sequences. Additionally, we used state of the art motion capture sequences with ground truth data for quantitative evaluation. 
The experiments showed that through incorporating temporal consistency on body shape and motion, our approach yields more accurate and robust results than SMPLify-X.
Moreover, we showed that our approach not only avoids jittering artifacts and inaccuracies, but also helps to produce plausible and realistic poses and smooth body motion.

\section{State of the Art}\label{sec:soa} 

    Fitting body model parameters to a monocular image sequence can be performed for every image individually, or with the remaining images of the sequence as context. 
    
    
    \subsection{Fitting a Body Model to Images} 
    
    
        Body model fitting is done on the basis of monocular camera systems, systems with multiple cameras, with depth sensors, or other configurations. The challenging task of fitting a body model to a single image without further depth information is discussed in the following.
        
        In order to capture whole body motions, a detailed representation of the human body is needed, such as meshes or parametric body models. Parameters of the body model can be fitted using keypoints of the image, like joint locations. SMPLify-X \cite{SMPLify-X} finds body shape and pose with articulated hands and facial expression for people pictured on a single RGB image. With OpenPose \cite{OpenPose}, 2D joint locations of the imaged people are estimated, to which the parameters of the camera and the body model SMPL-X \cite{SMPLify-X} are fitted. After an initialization phase for the camera parameters, the fitting is realized by projecting the model joints onto the image plane according to current camera properties and by minimizing the distance between projected joints and detected keypoints. At the same time, the deviation from the neutral state is penalized with pose and shape priors. 
        An interpenetration penalty is applied to suppress body parts intersecting each other. SMPLify-X's pose prior VPoser is realized as a variational autoencoder, which translates the body pose parameters into a normally distributed latent space. It is trained for the body model SMPL-X on a database of roughly one million recordings of common and uncommon body poses. SMPLify-X is the enhancement of SMPLify \cite{SMPLify} from the body model SMPL \cite{SMPL} to SMPL-X, which in contrast to SMPL includes articulated hands and facial expression.
        The research group of SMPLify-X thereafter published ExPose \cite{expose}, a body shape and pose estimation approach which achieves similar results to SMPLify-X, but is 200 times as fast \cite{expose}. SMPL-X parameters are regressed directly, similarly learned as in Human Mesh Recovery (HMR) \cite{HMR}, on a new SMPL-X dataset. The parameters for hands and facial expression are regressed separately by a network using the respective image parts in high resolution. Finally, the two parameter sets are merged and fine-tuned.
        
        Monocular full body capture in real-time is enabled by \cite{real-time}, which uses a neural network architecture consisting of residual, convolutional and fully connected modules and employs a combination of 2D and 3D keypoint positions and cropped parts of the input image for the face and hands as intermediate representation of the body.
        Inter-part correlations of hands and high-level body features are used to finetune hand and finger keypoint detection.
        Additionally to facial expression estimation, the cropped image is warped and projected onto the SMPL+H body model \cite{SMPL+H}.
        Alternatively, a feature pyramid with different image scales can be created from the input image to derive the shape of the depicted human. This is done by \cite{pyramidal}, employing an encoder which takes the image and produces a feature pyramid, supervised by pixel-wise image segmentation into body parts and background, on the highest resolution feature. 
        Then, there is a feedback loop over the feature pyramid, inside a regression network. There, model parameters of the body model SMPL \cite{SMPL} are estimated to mesh-image-alignment from features of increasing resolution. Due to the usage of SMPL as underlying body model, no facial expression or finger poses are estimated.
        
        Two prevailing methods for the stated problem are HMR \cite{HMR} and Unite the People \cite{Unite_the_people}, which both do not include articulated hands or facial expression. HMR learns a mapping from image pixels directly to SMPL parameters, from 2D keypoint annotated images and a large database of 3D human motion and shape, without paired data. It minimizes the reprojection loss of keypoints and uses a discriminator network. 
        In Unite the People, a database of paired data is generated with SMPLify \cite{SMPLify} to train a direct regression method.
        
        There are approaches for specialized problems, such as jointly identifying a person and the object they are interacting with \cite{PHOSA}, or specifying on low-resolution images \cite{LowRes2021}. 
        Camera systems with alternative capturing types, like polarized 2D images \cite{polar_images}, images with a depth channel \cite{rgbd}, or images from an event-based camera \cite{eventcap} 
        can be used to improve different aspects of body model parameter fitting based on individual images. 

    \subsection{Fitting a Body Model to Image Sequences}
        Analyzing an image sequence as a whole yields information about the motion, rather than the poses individually and thus about the temporal connection between the estimated poses.
        The Video Inference for Human Body Pose and Shape Estimation (VIBE) approach \cite{vibe} is based on using a motion discriminator for performing body shape and pose estimation on videos or image sequences. Because SMPL \cite{SMPL} is used as the underlying body model, it does not include articulated hands or facial expression. VIBE exploits the large-scale motion capture dataset AMASS \cite{AMASS} and in-the-wild, 2D keypoint annotated videos to train a motion discriminator model and estimate SMPL parameters. The motion discriminator is trained like a generative adversarial network, with two separate neural networks. 
        
        Similar to our 
        approach, Monocular Total Capture (MTC) \cite{MTC} captures body shape and pose, including expressive hands and face, of a person depicted in a monocular image sequence, represented by a 3D deformable mesh body model and uses one shared shape vector for the whole image sequence. For every image, the joint confidence maps and 3D orientation of body parts are estimated with a Convolutional Neural Network (CNN), which are described as 3D Part Orientation Fields (POFs). The POFs are used to optimize the parameters of the body model ADAM \cite{ADAM}, employing the pose and shape prior from the CMU Graphics Lab Motion Capture 
        Database \cite{Z_CMU_Database}. 
        For temporal consistency, a texture-based tracking method is utilized, using optical flow. It refines the body model parameters based on the previous image, its parameters, and the parameters of the current image.
        
        An approach concentrating on articulated hands on monocular videos is FrankMocap \cite{frank}. The poses and shapes of the hands are estimated separately from the body and are merged afterwards, as in ExPose. Body pose and hands are both modelled with SMPL-X, thus FrankMocap includes expressive face and articulated hands.
        Switching between methods to iteratively estimate shape and pose in an ongoing loop is an alternative approach for the given problem. SPIN \cite{SPIN} switches between the deep regression network of HMR and an iterative optimization routine, using SMPLify \cite{SMPLify}, to improve the accuracy of body shape and motion estimation. Due to choosing SMPL as the body model, SPIN does not include expressive face or articulated hands.
        
        
        For temporally consistent capturing of the human body pose and shape including clothes and texture, \cite{ClothedVideoHighRes} uses a cascaded 
        network architecture of image encoders and Multi-Layer Perceptrons (MLPs). First, a Temporal Voxel Regression Network (TVRN) produces temporally consistent voxel occupancy grids for every frame. Then, this is used by a geometry prediction network (H3DN) to refine the surface 3D reconstruction, by acquiring a multi-scale shape encoding. Afterwards, a texture prediction network (H3DTexN) is used to predict a color value for each surface vertex. Using a modified residual neural network (ResNet), pixel-wise image features are acquired, which are processed in an MLP decoder for every frame but with shared parameters, again acquiring a multi-scale shape encoding. This produces a textured 3D surface that captures finger pose and facial expression, but not in great detail.

        To capture humans in challenging poses and motion,  
        ChallenCap \cite{challencap} uses a learning-and-optimization framework, which learns motion characteristics and is trained on ``a new challenging human motion dataset with both unsynchronized marker-based and light-weight multi-image references'' \cite{challencap}.
        First, a noisy skeletal motion map is acquired, this is then processed in their temporal encoder-decoder and generation network HybridNet and their motion discriminator, which is followed by a robust motion optimization phase, which uses 2D keypoint and silhouette information from the video frames. This approach needs a 3D body scanned template mesh of the actor as input, besides the video frames, and does not capture facial expression or finger poses.
        
        Pure human pose estimation from image sequences or videos, outputting only a skeleton of joints instead of estimating the person’s body shape, is another challenging research area.
    A recent overview of deep learning based approaches to this task is provided by \cite{SurveyDL}. Almost 300 publications on 2D and 3D body pose estimation, related datasets and evaluation metrics are analyzed, discussed and compared.  In addition, remaining challenges such as occlusion, computational efficiency, and insufficient training data are identified.
    One of those publications is \cite{3DTempConv}, which performs pure 3D human pose estimation and uses a fully convolutional model, which consists of dilated temporal convolutions over 2D keypoints to capture long-term information. They propose to use this for applications where labeled data is scarce to enable semi-supervised learning on unlabeled video using a reprojection loss, similar to this paper.
    \cite{GraphCNNs} performs skeleton-based action recognition and adopts the possible dependencies of physically connected and disconnected joints and the hierarchical structure in the skeleton to better model the spatial and temporal features of human actions. It employs a spatial-temporal module which contains motif-based graph convolutional networks (GCNs), which work on graphs having human body joints as nodes and bones as edges, and which contain a variable temporal dense block to encode short-, mid and long-range local information. In the last stages of the network, a non-local neural network module for global temporal features is used, with the aid of an attention mechanism.
        
    \subsection{This Work's Contribution}
        
        This work presents a method for temporally consistent human body shape and motion estimation by fitting SMPL-X parameters to a monocular image sequence. It estimates 2D joint location keypoints bottom-up and then fits SMPL-X parameters to the keypoints top-down in an optimization framework. A consistent shape is found by 
        estimating one shared shape parameter vector for the whole image sequence.
        Following, the body motion is estimated accurately by imposition of temporal consistency. For this purpose, we minimize objective functions, which consist of a reprojection loss term, a number of pose and shape priors and loss terms for temporally consistent body shape and motion.

        Most similar to our 
        approach are VIBE \cite{vibe} and MTC \cite{MTC}. However, VIBE does not include expressive face and articulated hands and MTC does not focus on shape fitting and uses a gender non-specific body model, in contrast to our work. Therefore, our approach surpasses the other two in capturing people naturally and accurately, including detailed finger movements, emotional countenance, expressive gestures and a diverse spectrum of gender-specific body shape.
        In comparison to SMPLify-X, our approach exploits the context of temporally neighboring frames to provide body models which are temporally consistent in shape and motion.
        
        
        This enables the derivation of realistic 3D body models from image sequences, including facial expression and articulated hands. 
        We present extensive experiments and apply our approach to the challenging task of sign language motion capture.
      
\section{Temporally Consistent Shape and Pose Fitting}\label{sec:method}
Our approach takes an image sequence of a single person in motion as input and generates a sequence of body models with a consistent shape and smooth motion of the imaged person, including facial expression and articulated hands. 

\begin{figure*}[tb]
    \centering
    \includegraphics[width=\linewidth]{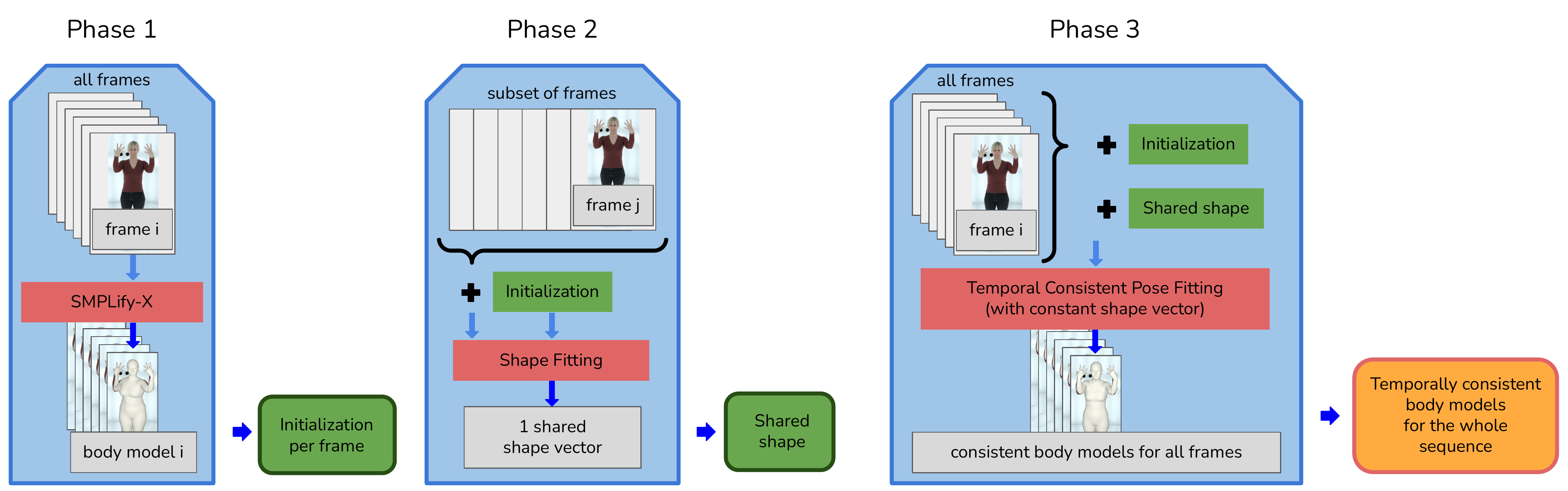}
    \caption[Method Overview]{Method overview with three phases, including interim results as the green blocks and optimization frameworks as the red blocks. Latter all contain a reprojection error to OpenPose keypoints and assorted priors, consist of 5 stages with different weights, while phase 1 optimizes on individual frames and phase 2 and 3 on sets of frames.}
    \label{fig:method}
\end{figure*}

As shown in the method overview in Fig.~\ref{fig:method}, body model parameters are fitted to the image sequence in three phases. In phase 1, the image sequence is processed consecutively with OpenPose \cite{OpenPose} to obtain predicted joint locations as keypoints at image pixel positions per image. These keypoints are fed into SMPLify-X \cite{SMPLify-X} to calculate an initialization for the desired temporally consistent body shape and motion (Sec.~\ref{sec:init}). In phase 2, a consistent shape for the image sequence is derived from the initialized body models by optimization (Sec.~\ref{sec:shapefitting}). 
Finally, in phase 3, the SMPL-X parameters are optimized along the sequence for temporally consistent body motion (Sec.~\ref{sec:timecons}).

We use SMPL-X \cite{SMPLify-X} as underlying body model, because it is particularly well suited for 
the capturing of expressive motion 
since it includes facial expression and articulated hands. SMPL-X uses the following parameters: human body shape, general pose, finger poses, jaw pose, facial expression, and global rotation of the model.
    Finally, the camera center
    is also a parameter of interest for our 
    approach. 
 
    In all three phases of our optimization framework, the respective objective function includes the objective function from SMPLify-X, denoted by the loss term $E_{o_i}$, which is further described in Sec.~\ref{sec:init}.
    In phase 2, additionally, the consistent shape fitting loss $E_{\bm{\beta'}}$ is optimized, while phase 3 also includes the temporal consistency loss $E_{T_i}$, introduced in Sec.~\ref{sec:shapefitting} and~\ref{sec:timecons} respectively.

\subsection{Phase 1: Initialization}\label{sec:init}

   We use conventional SMPLify-X results for the given image sequence as initialization for phases 2 and 3 of the temporally consistent shape and pose fitting.
    We generate keypoints for the images using OpenPose and feed them to SMPLify-X for the estimation of an initial body model and camera parameters. The optimization of SMPLify-X's objective function for each image of a sequence $\mathcal{I}$ can be achieved by optimizing the following objective function (Eq.~\ref{eq:overall_smplx}).
    \begin{equation} 
    \label{eq:overall_smplx}
    E_1(\text{B},\Theta,\Psi,\bm{\lambda}) =\frac{1}{|\mathcal{I}|} \sum_{i \in \mathcal{I}} E_{o_i}
    \end{equation}
    The set of images used in the current optimization step is denoted as $\mathcal{I}$.
    From here on, all capitalized Greek letters are matrices of dimension $d \times |\mathcal{I}|$, containing the respective body model parameter vector (with dimension $d$) for every image in the sequence. For example $\text{B}$ is the matrix $(\bm{\beta}_1, ..., \bm{\beta}_{|\mathcal{I}|})$, where $\bm{\beta}_i$ is the shape parameter vector of the body model belonging to image $i$. $\Theta$ is the matrix of body pose parameters for every image, including finger pose $\Theta_{h}$, jaw pose $\Theta_{f}$ and general body pose $\Theta_{b}$, while the parameters for facial expression are described by $\Psi$. Finally, $\bm{\lambda}$ describes weights used from SMPLify-X that steer the influence of each term inside $E_{o_i}$.
    
    SMPLify-X's objective function, here described with slightly different notation by the loss term $E_{o_i}$ for image $i$, implements priors for jaw pose, hand poses, body shape, and facial expression, denoted by $E_{\bm{\theta}_{f_i}}$, $E_{\bm{\theta}_{h_i}}$, $E_{\bm{\beta}_i}$ and $E_{\bm{\psi}_i}$ respectively, which are all defined as $\ell^2$-norms. A reprojection loss, further explained below, is used as data term $E_{J_i}$ for image $i$.
    As body pose prior $E_{\bm{\theta}_{b_i}}$, SMPLify-X's prior VPoser is used, a deep learning based method discussed in Sec.~\ref{sec:soa}. A prior against overbending of knees and elbows $E_{\alpha_i}$ is deployed, penalizing the deviation from 0 of those body pose parameters exponentially. 
        The prior $E_{\mathcal{C}_i}$ avoids collisions between body parts by finding colliding faces of the mesh employing Bounding Volume Hierarchies \cite{bvh} and penalizing these by the depth of intrusion. 
        \begin{align} \label{eq:o_i}
        E_{o_i}&(\text{B},\Theta,\Psi,\bm{\lambda}) = \lambda_{J} E_{J_i} + \lambda_{\Theta_b} E_{\bm{\theta}_{b_i}} + \lambda_{\Theta_f} E_{\bm{\theta}_{f_i}}  \nonumber\\ &+ \lambda_{\Theta_h} E_{\bm{\theta}_{h_i}} + \lambda_\alpha E_{\alpha_i}
        + \lambda_\text{B} E_{\bm{\beta}_i} + \lambda_{\Psi} E_{\bm{\psi}_i} + \lambda_{\mathcal{C}} E_{\mathcal{C}_i}
        \end{align}
        The different $\bm{\lambda}$s weigh the five optimization stages SMPLify-X undergoes, which are explained below. Further details on these priors are discussed in SMPLify-X \cite{SMPLify-X}. 
        
\begin{figure}[tb]
            \centering
            \includegraphics[width=\linewidth]{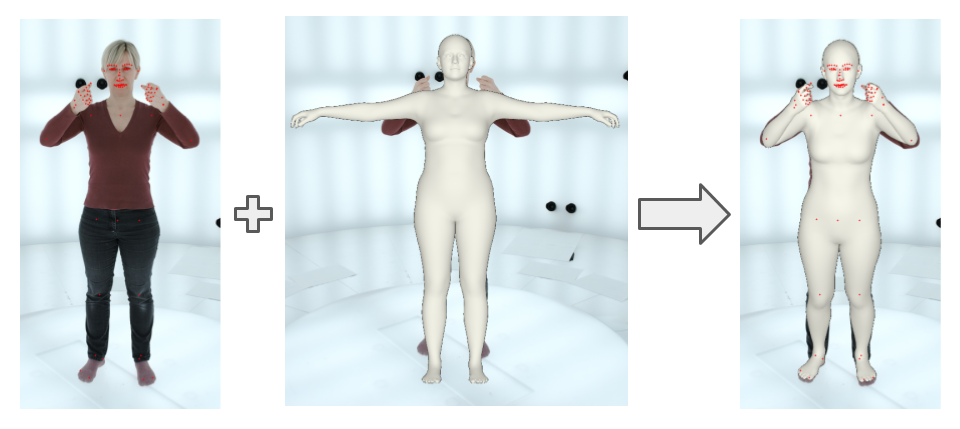}
            \caption{Illustration of reprojection loss: Input image with OpenPose keypoints marked in red (left), SMPL-X model in initial rest-pose (center) and pose and shape adapted SMPL-X model according to the OpenPose keypoints (right).}
            \label{fig:repr_loss}
        \end{figure}

    
        The reprojection loss (example in Fig.~\ref{fig:repr_loss}) inside the data term $E_{J_i}$ 
        is computed such that for each SMPL-X body joint projected to the image plane, a penalty value obtained from the distance to the corresponding OpenPose keypoint is summed up. 
        
        More precisely, the data term containing the reprojection loss is:
        \begin{equation}
        \begin{split}
        E_{J_i}(\text{B},&\Theta,K,J_\textit{est}) =\\
        &\sum_{\textit{joint }j} \gamma_j \omega_j \rho(\Pi_K(R_{\bm{\theta}_i}(J(\bm{\beta}_i))_j) - J_\textit{est,j})
        \end{split}\label{eq:repr_loss}
        \end{equation} 
        As described in \cite{SMPLify-X}, this loss minimizes the distance between the 2D keypoints $J_\textit{est}$ and the 2D projections of the corresponding  3D joints $R_{\bm{\theta}}(J(\bm{\beta}))_j$. $R_{\bm{\theta}}(\cdot)$ describes a function that transforms the joints along the kinematic chain according to the pose $\bm{\theta}$. $\rho$ denotes a robust error function to account for noisy detections and $\gamma_j$ are per-joint weights for annealed optimization. 
        
    In SMPLify-X's approach, the first step is the initialization of the depth of the camera based on the keypoint locations for the shoulders and hips estimated for the given image. Then, an iterative fitting begins, consisting of five stages in which the camera and SMPL-X model parameters are fitted to the given keypoints for the image. In every stage, the presented objective function is optimized.
    
    SMPLify-X's weights $\bm{\lambda}$ (given in Tab.~\ref{tab:gewichtung_loss}) are configured in such a way that
    at the beginning of the five stages, primarily body model parameters are strongly regulated with priors to achieve realistic human motion. As the stages progress, priors for jaw pose and inhibiting overbending of knees and elbows gain weight in relation to the rest. 
    Eventually, the weighting focus shifts to minimizing the reprojection loss and to details such as hand poses, facial expression, and collision prevention between body parts.
    \begin{table}[htb]
        \centering
        \begin{tabular}{l c|c|c|c|c} 
            \multicolumn{1}{c }{\textbf{Stage}} & \textbf{1} & \textbf{2} & \textbf{3} & \textbf{4} & \textbf{5} \\
            \hline
            \hline
            \hline 
            $\lambda_{\Theta_b}$ body pose prior & 404 & 404 & 57.4 & 4.8 & 4.8 \\ 
            $\lambda_{\Theta_h}$ hand pose prior & 404 & 404 & 57.4 & 4.8 & 4.8 \\ 
            $\lambda_\text{B}$ shape prior & 100 & 50 & 10 & 5 & 5 \\ 
            $\lambda_{\Psi}$ expression prior & 100 & 50 & 10 & 5 & 5 \\ 
            $\lambda_{\Theta_f}$ jaw pose prior 1 & 63.6 & 63.6 & 24 & 6.9 & 6.9 \\ 
            $\lambda_{\Theta_f}$ jaw pose prior 2 & 201 & 201 & 75.7 & 21.9 & 21.9 \\ 
            $\lambda_{\Theta_f}$ jaw pose prior 3 & 201 & 201 & 75.8 & 21.9 & 21.9 \\ 
            $\lambda_\alpha$ bending prior & 35.8 & 35.8 & 13.5 & 3.9 & 3.9 \\ 
            $\lambda_{\mathcal{C}}$ collision loss & 0 & 0 & 0 & 0.1 & 1 \\ 
            $\lambda_{J}$ hand joints & 0 & 0 & 0 & 0.1 & 2 \\ 
            $\lambda_{J}$ face joints & 0 & 0 & 0 & 0 & 2 \\ 
            $\lambda_{J}$ rest of joints & 1 & 1 & 1 & 1 & 1 
        \end{tabular}
        \caption{Weights $\bm{\lambda}$ used by SMPLify-X and the following phases for the different priors and subterms in Eq.~\ref{eq:o_i}.}
        \label{tab:gewichtung_loss}
    \end{table}
    
    For optimization of the objective function, SMPLify-X uses the limited-memory BFGS (L-BFGS) algorithm \cite{bfgs}.
    All body model parameters, the camera center and the global rotation undergo optimization in this phase.
    The result of this phase are temporally inconsistent shape and pose estimations from SMPLify-X, which are used as initialization for the next two phases.

\subsection{Phase 2: Shape Fitting on Subset of Images}\label{sec:shapefitting}
    
    Our goal is to estimate one consistent body shape parameter vector for the whole image sequence. As the optimization of a full sequence is computationally too expensive, we optimize over a subset $\mathcal{I}_{\textit{sub}} \subset \mathcal{I}$.
    The shape parameters are derived solely from the estimated keypoints from OpenPose and their proportional relations. 
    
    This phase is similar to the first phase, despite optimizing one shared shape over a subset of images instead of individual ones over all images. 
    This is achieved by optimizing objective function $E_2$:
    \begin{equation} 
    \label{eq:overall_shape}
    E_2(\text{B},\Theta,\Psi,\bm{\lambda}) = E_{\bm{\beta}'} +\frac{1}{|\mathcal{I}_{\textit{sub}}|} \sum_{i \in \mathcal{I}_{\textit{sub}}} E_{o_i}
    \end{equation}
    As the SMPLify-X loss term $E_{o_i}$ and the variables are the same as explained for objective function $E_1$ of phase 1 (Eq.~\ref{eq:overall_smplx}) and $\bm{\beta'}$ describes the shared shape parameter vector for the whole image sequence, a shape prior $E_{\bm{\beta'}}$ for the consistent shape is employed, which regularizes the shared shape parameter vector to a neutral shape and is realized as a weighted $\ell^2$-norm:
\begin{equation}
    E_{\bm{\beta'}}(\bm{\beta'})={w}_{\bm{\beta'}} \cdot \sum_{i=1}^n x_i^2
\end{equation} 
    with $\bm{\beta'}=[x_1, ..., x_n]^\top$ and ${w}_{\bm{\beta'}}$ being the shape prior weight for the shared shape parameter vector. 

    As initialization for this phase's optimization step, the results from phase 1 (Sec.~\ref{sec:init}) is used, but the individual shape vectors per image is replaced by the shared shape parameter vector $\bm{\beta'}$ as estimated shape for every image of the sequence. This shared shape vector is initialized with the mean of the neglected shape vectors. More formally put: Let $M_{\textit{sub}}$ be the set of body models for the images in $\mathcal{I}_{\textit{sub}}$, which were found in phase 1 (Sec.~\ref{sec:init}). As initialization for this phase, $M_{\textit{sub}}$ is updated to the set $\hat{M}_{\textit{sub}}$ by changing the shape parameter matrix 
    $\text{B}$ to $[\bm{\beta'}, \text{ ... } , \bm{\beta'}]$
    for this phase, so that the shared shape parameter vector is used as estimated shape for every image of the sequence. 
    Then, $\bm{\beta'}$ is initialized with the mean of the shape parameters of the body models in $M_{\textit{sub}}$.

    Optimization of the objective function of this phase (Eq.~\ref{eq:overall_shape}) is performed as described in phase 1, using the L-BFGS algorithm. The shared shape parameter vector $\bm{\beta'}$, all body model parameters except the independent shape parameter vectors, the camera center and the global rotation undergo optimization here.
    It undergoes five stages of optimization, with the same weights $\bm{\lambda}$ as in phase 1.
    The weight ${w}_{\bm{\beta'}}$ for the shared shape prior $E_{\bm{\beta'}}$ is set to the values of the shape prior weight $\lambda_\text{B}$ from phase 1 (Sec.~\ref{sec:init}).
    
    
    
    The subset $\mathcal{I}_{\textit{sub}}$ is sampled randomly from $\mathcal{I}$. Alternative choices for $\mathcal{I}_{\textit{sub}}$ are possible, but taking random images was found to be sufficient, because in the case of inaccurate keypoints in a sub-sequence of the image sequence the random sample most likely also contains keypoints for different parts of the image sequence.
    Another possibility is to choose the $|\mathcal{I}_{\textit{sub}}|$ images with the highest average keypoint confidence score. 
    To obtain complete body shape information, such a set of images could alternatively be chosen such that no part of the body remains hidden in the course of the images.
    
    The result of the shared shape parameter $\bm{\beta'}$ is saved and is set constant for phase 3 which estimates a smooth motion over time (Sec.~\ref{sec:timecons}).

\subsection{Phase 3: Temporal Consistency on all Images}\label{sec:timecons}



    The goal of this phase is to estimate camera and body pose parameters for the entire input image sequence, which describe temporally consistent body motion,
    while the shape parameters are set to the consistent shape $\bm{\beta'}$ for every image $i$, which was found in phase 2 (Sec.~\ref{sec:shapefitting}), and are not considered for optimization anymore.
    Therefore, the objective function $E_3$ of this phase includes a temporal consistency loss $E_{T_i}$, the SMPLify-X loss term $E_{o_i}$ and the variables as explained for objective function $E_1$ of phase 1 (Eq.~\ref{eq:overall_smplx}):
    \begin{equation} 
    \label{eq:overall_temp}
    E_3(\text{B},\Theta,\Psi,\bm{\lambda}) = \frac{1}{|\mathcal{I}|} \sum_{i \in \mathcal{I}} \big(E_{o_i}+  E_{T_i}\big) 
    \end{equation}

\begin{figure}[ht]
        \centering
        \includegraphics[width=\linewidth]{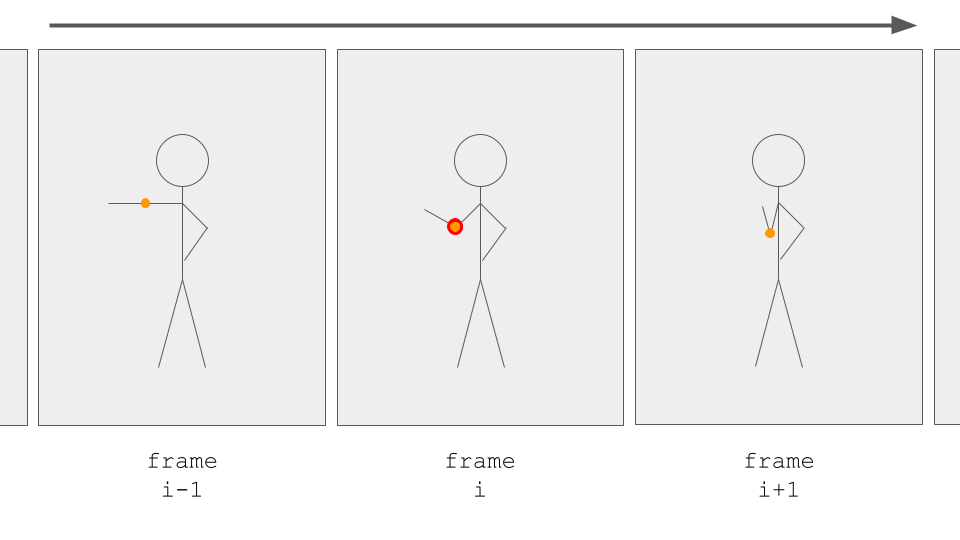}
        \caption{Temporal consistency calculation (Eq.~\ref{eq:temp_cons}) for frame i, for the elbow joint marked in orange and the temporal mean position of the joint in red.} 
        \label{fig:temp_loss}
    \end{figure}

    For smooth body motion all body parts follow a steady individual path over the course of the image sequence.
    Therefore, the underlying assumption used in this phase is that the 3D position of every body part is close to the mean position of that body part of the temporally neighboring frames. 
    Fig.~\ref{fig:temp_loss} illustrates this approach, where the mean position of the elbow joint in the neighboring frames is identical to the position in frame i, hence the temporal consistency loss is 0 for this very frame and joint. 

    The term $E_{T_i}$ describes the temporal consistency loss for the body model corresponding to image $i$, by calculating the distance between 3D joint positions of the body model of the current frame and of the mean of the temporal predecessor, current and successor frames.
    \begin{equation} 
        \begin{split}
            &E_{T_i}(\bm{\beta'},\Theta) = \\
            &\begin{cases}
                {w}_T \sum_{joint j}\Big(R_{\bm{\theta}_{i}}\big(J({\bm{\beta'}})\big)_j - m_{i,j}\Big)^2      & \text{if } 2 \leqslant i \leqslant |\mathcal{I}|-1,\\
                0                   & \text{else,}      \label{eq:temp_cons}
            \end{cases}
        \end{split}
    \end{equation}
    with
    \begin{align}
        m&_{i,j} = \nonumber \\
            & \frac{1}{3}\Big(R_{\bm{\theta}_{i-1}}\big(J({\bm{\beta'}})\big)_j + R_{\bm{\theta}_{i}}\big(J({\bm{\beta'}})\big)_j + R_{\bm{\theta}_{i+1}}\big(J({\bm{\beta'}})\big)_j\Big)                 \label{eq:mean}
    \end{align}
    The first case of Eq.~\ref{eq:temp_cons} defines the temporal consistency loss term for image frame 2 to the second last frame of image sequence $\mathcal{I}$. The second case ensures that, for the first (resp. last) frame of $\mathcal{I}$ the temporal consistency loss-value is 0, which is because this frame does not have a predecessor (resp. successor) frame.
    Following the notation of \cite{SMPLify-X}, $R_{\bm{\theta}_i}(J({\bm{\beta'}}))_j$ is the 3D position of joint $j$ of the posed and shaped body model belonging to image $i$, according pose parameter vector $\bm{\theta}_i$ and shared shape parameter vector $\bm{\beta'}$.
    Therefore, $m_{i,j}$ is the mean position of joint $j$ over the neighboring images of image $i$.
    To tune to what extent temporal consistency is imposed onto the body model sequence, the temporal consistency weight ${w}_T$ can be used.
    $E_{T_i}$ partially resembles the data term $E_{J_i}$ from SMPLify-X, where the 2D joint positions of the current shaped and posed body model are set against the 2D keypoints, while here, the 3D joint positions of the current shaped and posed body model are set against the mean 3D position of the joints of the frames in the temporal neighborhood.

    
    All body models are initialized with the pose estimates of phase 1 (Sec.~\ref{sec:init}) while the shape parameter vectors are set to the consistent shape $\bm{\beta'}$ found in phase 2 (Sec.~\ref{sec:shapefitting}). 
    The weights $\bm{\lambda}$ inside $E_{o_i}$ for the five stages of optimization are set as described in phase 1 (Sec.~\ref{sec:init}).
    The body model parameters for pose, the global rotation and the camera center for each image $i$ undergo optimization.
    
    Instead of averaging the joint positions over time for a measure of temporal consistency, other approaches are possible, like averaging the joint angles or averaging the positions over more than two frames. 
    
    Optimizing the objective function of this phase (Eq.~\ref{eq:overall_temp}) in the five-stage manner from phase 1 and 2 for the entire potentially long image sequence at once, is computationally complex for common optimizers.
    To efficiently approximate this, a Sliding Window Algorithm is introduced, which is described in the following.\\

\textbf{Optimization with the Sliding Window Algorithm:} 
    The presented objective function of phase 3 (Eq.~\ref{eq:overall_temp}) ensures temporal consistency of body poses for a complete image sequence. 
    The increasing number of parameters undergoing optimization for longer image sequences leads to a drastic increase in computational complexity for common optimizers, which increases the runtime. Therefore, this section presents a method approximating the optimum, using a sliding window approach.
    Instead of optimizing the body model parameters of all images in a given image sequence at once, it optimizes the body model parameters of all subsets of consecutive images of a given window size $n$, starting at the first $n$ images, iterating through the image sequence.
    
\begin{figure}[tbh]
        \centering
        \includegraphics[width=\linewidth]{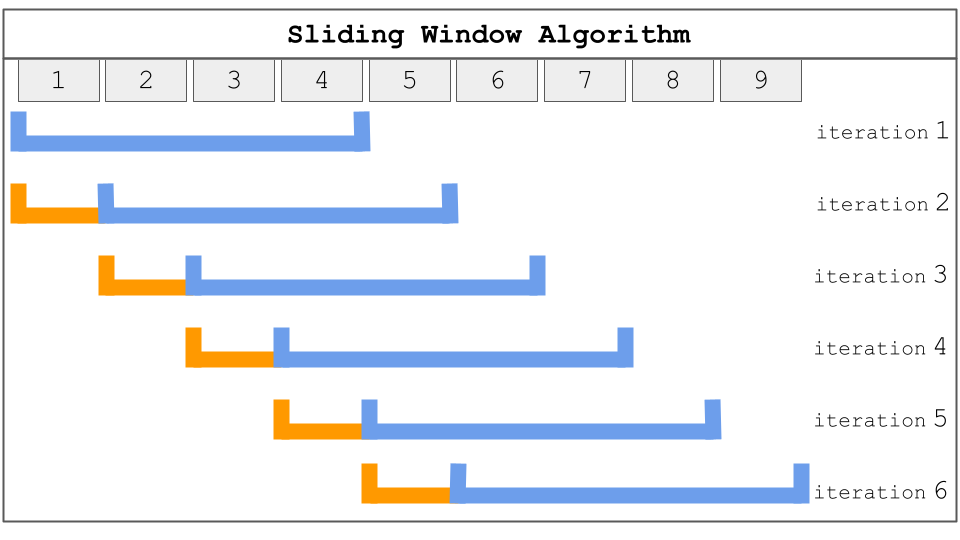}
        \caption[Procedure of the Sliding Window Algorithm]{Procedure of the Sliding Window Algorithm over image sequence of 9 frames (marked in gray) with window size 4. The frames underlined in blue denote the frames inside the sliding window, which currently get optimized, while the orange underlined frames are not optimized (body model parameters set to constant) but influence the optimization, because the temporal loss term (Eq.~\ref{eq:temp_cons}) is calculated for all frames in the window except the last.}
        \label{fig:sliding_window}
    \end{figure}

    In each iteration the presented five-stage optimization is performed on these $n$ images, optimizing their body model parameters, minimizing the objective function  $E_3$ (Eq.~\ref{eq:overall_temp}). Since the temporal consistency term of $E_3$ for a frame uses the mean joint position of the temporally neighboring frames, the calculation of this term for the last frame in each window would also use the frame following the window and the associated body model. However, because this body model is not optimized until the next iteration, only the rough initialization from phase 1 would be used for this optimization. Therefore, the temporal consistency term is not calculated for the last frame in the window, because this would lead to more inaccurate results. An illustration is given in Fig.~\ref{fig:sliding_window}.

    The result for image $i$ of the present iteration is final if this was the last iteration in which the image was inside the sliding window. Otherwise it functions as a good initialization for the upcoming iteration. 
    A smaller window size leads to a less accurate approximation of the objective function (Eq.~\ref{eq:overall_temp}). 
    %
    The temporally consistent motion enforced in this phase 3, together with the unified shape in phase 2, lead to a temporally consistent body shape and motion sequence by optimizing the objective function $E_3$ (Eq.~\ref{eq:overall_temp}) with the presented Sliding Window Algorithm. The performance of the presented approach is analyzed and compared to SMPLify-X in the following section by qualitative and quantitative evaluation.

\section{Evaluation}\label{sec:evaluation} 
    
    In the following, it is analyzed if the application of temporally consistent shape and pose fitting to an image sequence as described in Sec.~\ref{sec:method} produces an accurately fitted body model per image and smooth estimated motion, including expressive face and hands. 
    
    This is evaluated qualitatively by projecting the estimated body models onto the images according to the estimated camera parameters. These projections are compared to the original images and to the projections of the body models which were found by applying SMPLify-X \cite{SMPLify-X} for every image.
    Quantitative evaluation includes body shape evaluation by comparing volumes of the estimated body models to volumes of ground truth meshes
    , which were captured using a marker-based motion capture system.
    To furthermore measure the performance of pose estimation, the mean vertex-to-surface error to the ground truth meshes is analyzed.
    
    People shown in image sequences, whose body pose and shape estimates are evaluated in this section, are called \textit{subjects} from here on.
    
\subsection{Setting}\label{sec:eval_setting}
    
    \textbf{Architecture:} All following computations were performed on a system with two 20 core Xeon CPUs, 128GB RAM and four NVIDIA RTX 2080 GPUs (12GB VRAM each). OpenPose was executed in parallel on all four GPUs, while SMPLify-X and the algorithms developed in our work were not, which enabled processing four image sequences at the same time.
    
    \textbf{Datasets:} To evaluate our 
    approach, its performance was analyzed and compared on databases from two sources. 
    The first database is our own image sequence, consisting of 722 images, from here on called own-data.
    In this, the subject uses sign language and is captured from the front.
    
    The second database consists of videos from MoVi \cite{MoVi}, which were processed to image sequences. MoVi is a large marker-based motion capture dataset on videos with actors wearing natural or minimal clothing and performing various body motions. The body model and camera parameters found by MoVi were converted to SMPL+H \cite{SMPL+H} and additional soft-tissue deformations parameters by AMASS \cite{AMASS}. Therefore this database is well suited for quantitative evaluation in this paper. 
    
    From MoVi, we chose six subjects for evaluation and the corresponding body models from AMASS are used as ground truth, meaning these models are assumed to be a good representation of the real subjects and their poses and serve as reference for the evaluation phase.
    The three subjects 17, 1 and 32 have average body volume, while subjects 41 and 16 have higher and subject 45 has lower body volume than average.
    In the videos of the motion capture dataset, all subjects are depicted from a slight side view and perform the same 21 body motions. Two of these motions were selected for further quantitative and qualitative evaluation in our work. Hand waving was chosen as a simple motion for body shape and pose estimation, which is from here on referred to as \textit{Simple Motion}, and sitting down cross-legged and standing up again as a difficult one, which is from here on called \textit{Complex Motion}. The MoVi video sequences were processed to image sequences with the original framerate of 25 images per second, which results in an individual number of frames for every subject and motion, given in Table~\ref{tab:frame_numbers}. In some instances for these subjects, MoVi starts the motion capture process slightly later inside the video or stops it slightly earlier than our work started and ended the image sequences, thereby some frames at the end or beginning of the image sequence have no corresponding ground truth result.
    
    \begin{table}[ht]
        \centering
        \begin{tabular}{c|c|c}
           \makecell{Motion sequence} & \makecell{Total number \\ of frames} & \makecell{Number of \\ frames with \\ ground truth}\\
           \hline
           \hline
           \hline
           \textbf{Simple Motion} & & \\
           subject 1 & 50 & 50 \\
           subject 16 & 75 & 75 \\
           subject 17 & 75 & 75 \\
           subject 32 & 100 & 85 \\
           subject 41 & 100 & 97 \\
           subject 45 & 75 & 62 \\
           \hline
           \textbf{Complex Motion} & & \\
           subject 1 & 225 & 222 \\
           subject 16 & 200 & 182 \\
           subject 17 & 175 & 162 \\
           subject 32 & 250 & 247 \\
           subject 41 & 219 & 219 \\
           subject 45 & 177 & 175 \\
        \end{tabular}
        \caption[Image sequence frame numbers from MoVi videos]{The number of frames for every image sequence taken from the complete MoVi video of a subject. In some instances ground truth data was not available for frames from the beginning or end of a sequence.}
        \label{tab:frame_numbers}
    \end{table}
    
    All following experiments use the parameter settings given in Table~\ref{tab:eval_settings}, unless otherwise specified.
    
    \begin{table}[ht]
        \centering
        \begin{tabular}{c|c}
            \textbf{Setting} & \textbf{Value} \\
            \hline
            \hline
            \hline
            temporal consistency weight & 100 \\
            shape sample size & 15 \\
            window size & 7 \\
        \end{tabular}
        \caption[Common parameter settings of subsequent experiments]{Common parameter settings of subsequent experiments.}
        \label{tab:eval_settings}
    \end{table}

\subsection{Shape Fitting Evaluation}

            
            
        
        In this section, the shape fitting results of our  approach are evaluated and compared to SMPLify-X \cite{SMPLify-X} and VIBE \cite{vibe}. 
        
\subsubsection{Qualitative Analysis}
        
\begin{figure}[t]
              \centering
                \begin{subfigure}[t]{\linewidth}
                    \centering
                     \includegraphics[width=0.19\textwidth]{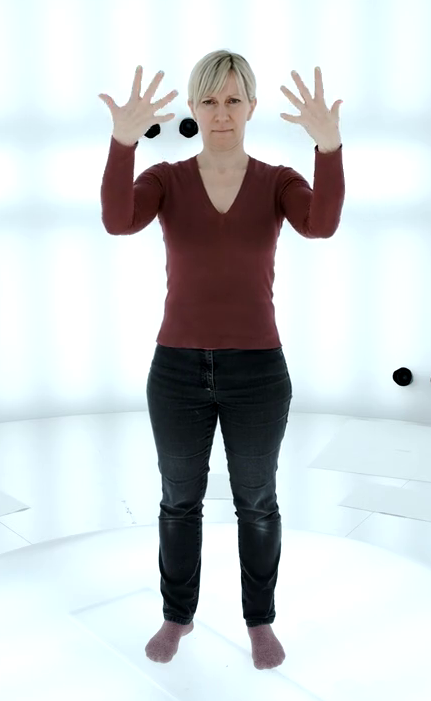}
                    \includegraphics[width=0.19\textwidth]{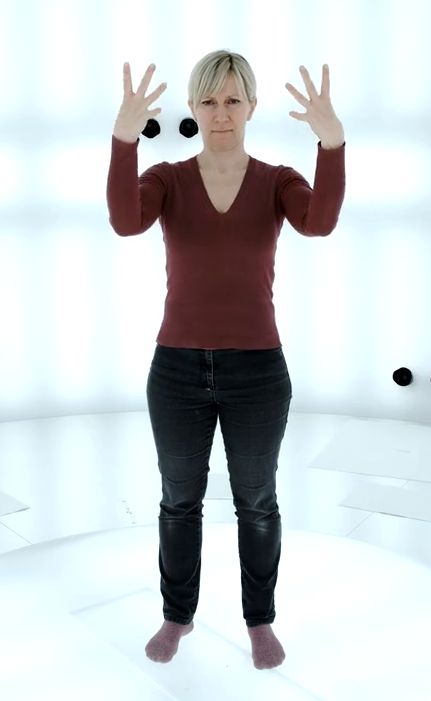}
                    \includegraphics[width=0.19\textwidth]{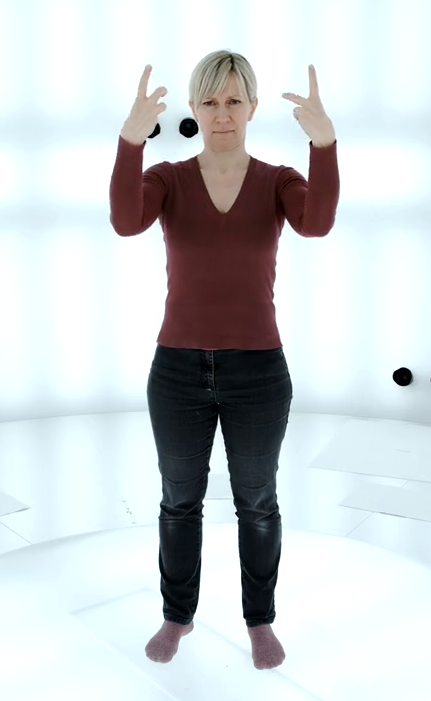}
                    \includegraphics[width=0.19\textwidth]{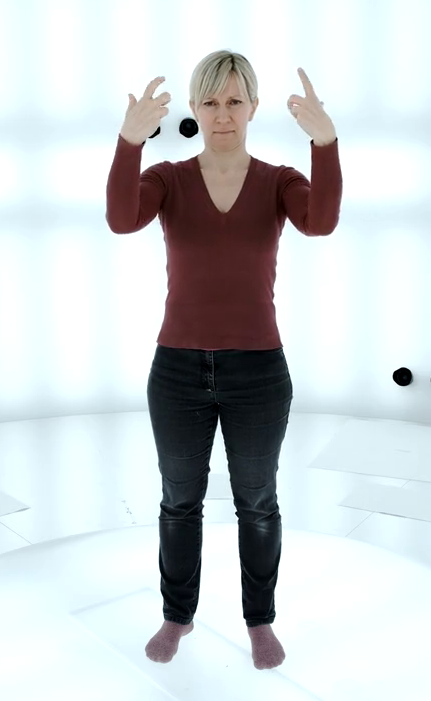}
                    \includegraphics[width=0.19\textwidth]{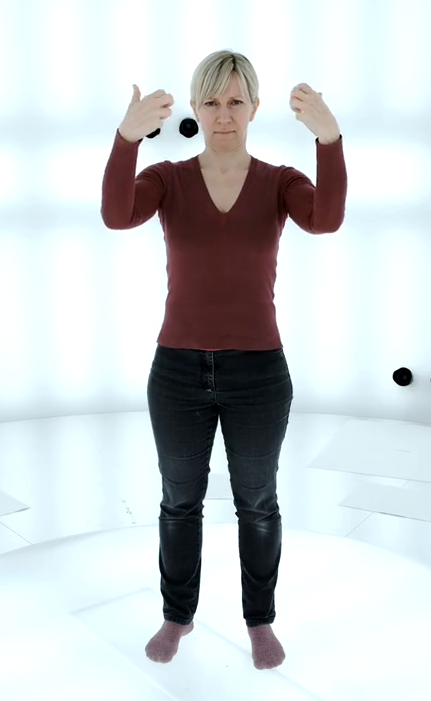}
                    \caption{Original frames}
                        \vspace*{4mm}
                \end{subfigure}
                
                \begin{subfigure}[t]{\linewidth}
                    \centering
                     \includegraphics[width=0.19\textwidth]{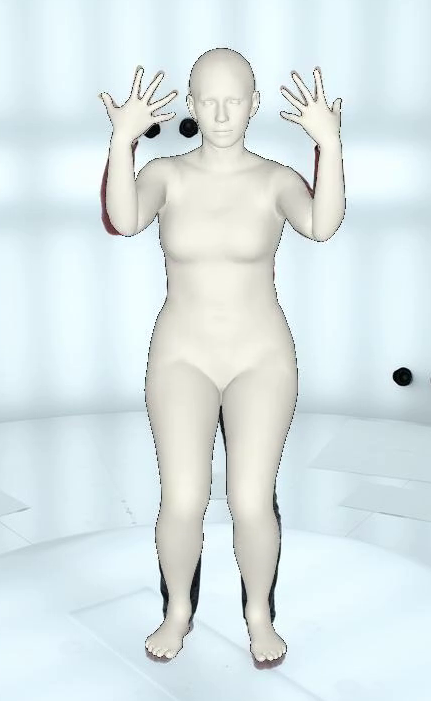}
                    \includegraphics[width=0.19\textwidth]{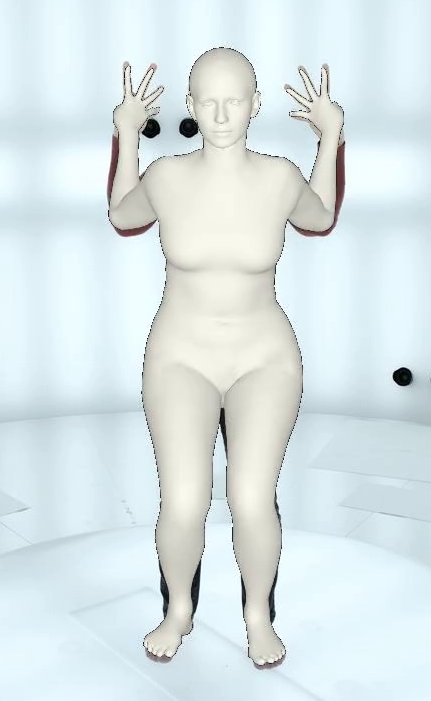}
                    \includegraphics[width=0.19\textwidth]{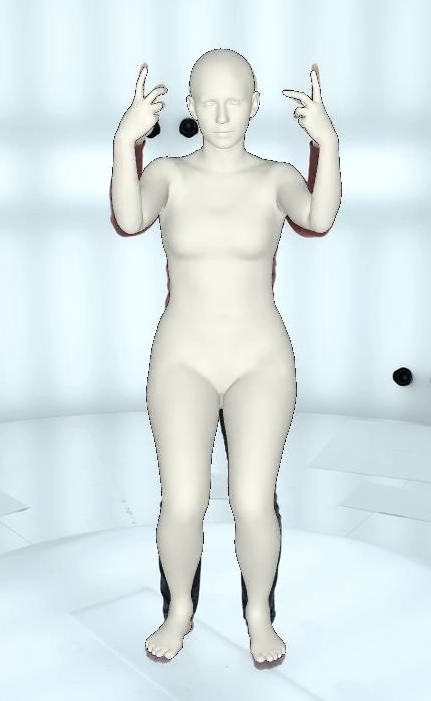}
                    \includegraphics[width=0.19\textwidth]{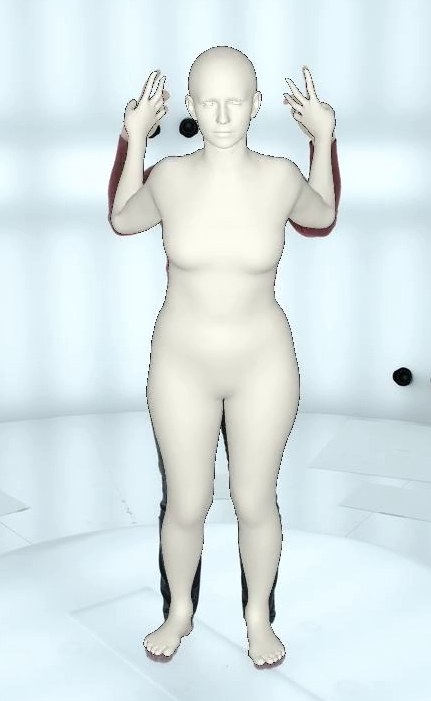}
                    \includegraphics[width=0.19\textwidth]{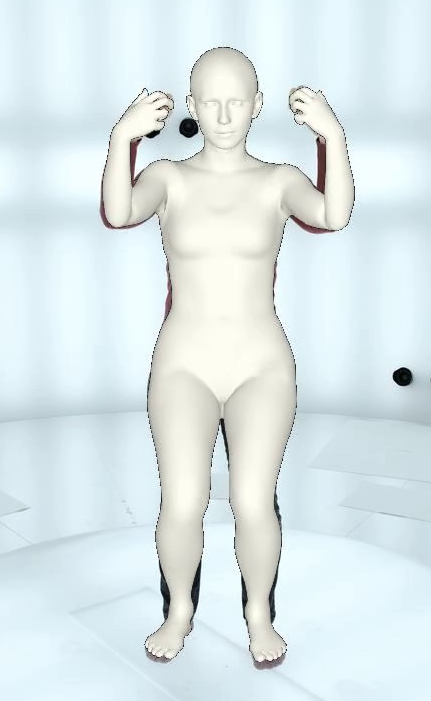}
                    \caption{SMPLify-X results}
                        \vspace*{4mm}
                \end{subfigure}

                \begin{subfigure}[t]{\linewidth}
                    \centering
                     \includegraphics[width=0.19\textwidth]{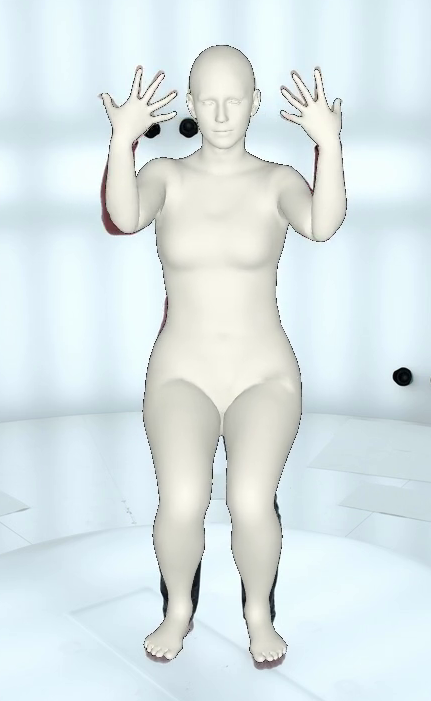}
                    \includegraphics[width=0.19\textwidth]{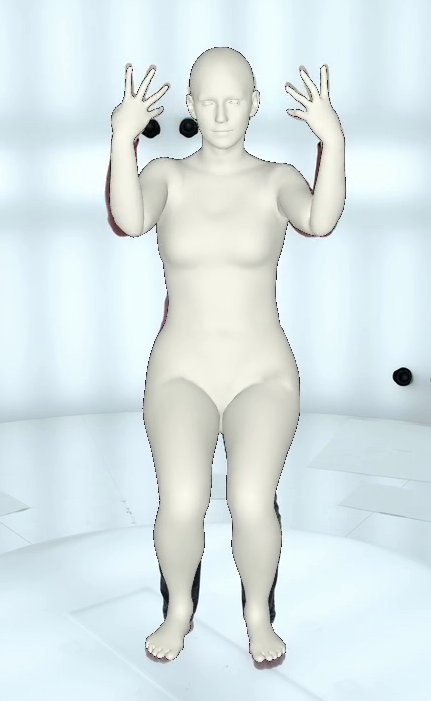}
                    \includegraphics[width=0.19\textwidth]{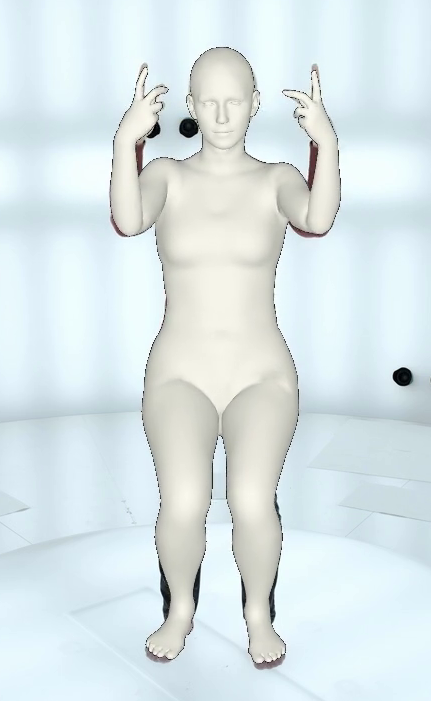}
                    \includegraphics[width=0.19\textwidth]{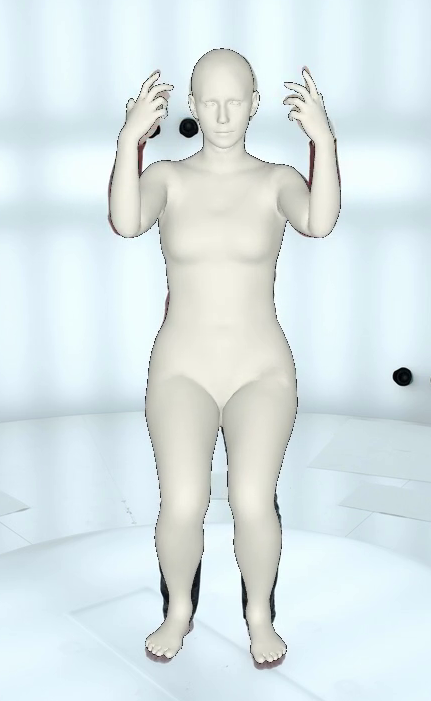}
                    \includegraphics[width=0.19\textwidth]{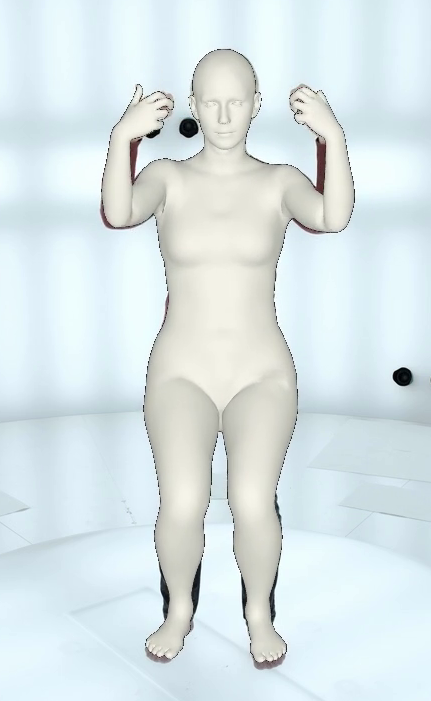}
                    \caption{Our results}
                \end{subfigure}
                
                \begin{subfigure}[t]{\linewidth}
                    \centering
                     \includegraphics[width=0.19\textwidth]{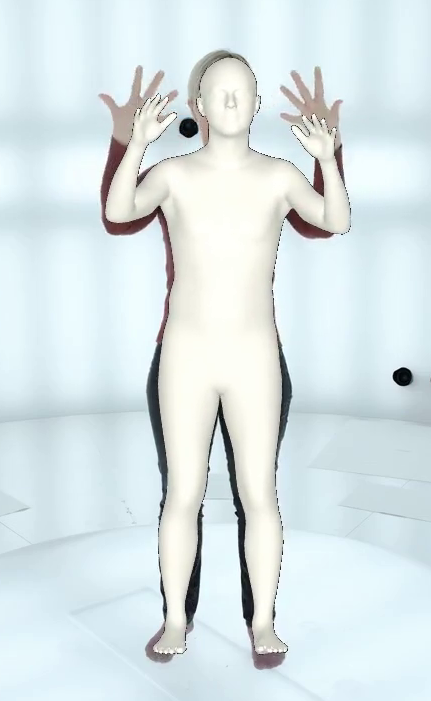}
                    \includegraphics[width=0.19\textwidth]{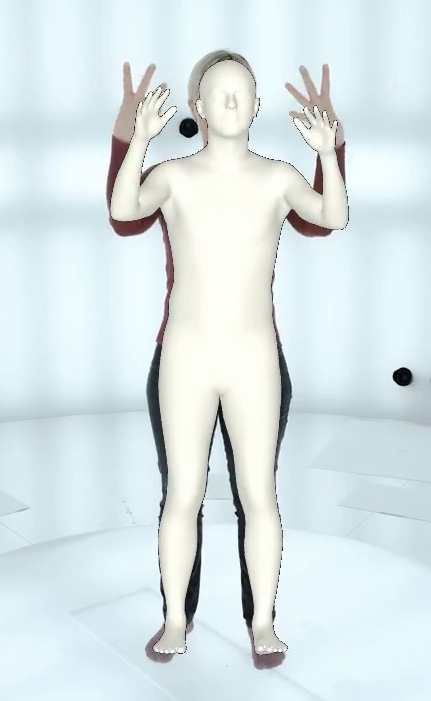}
                    \includegraphics[width=0.19\textwidth]{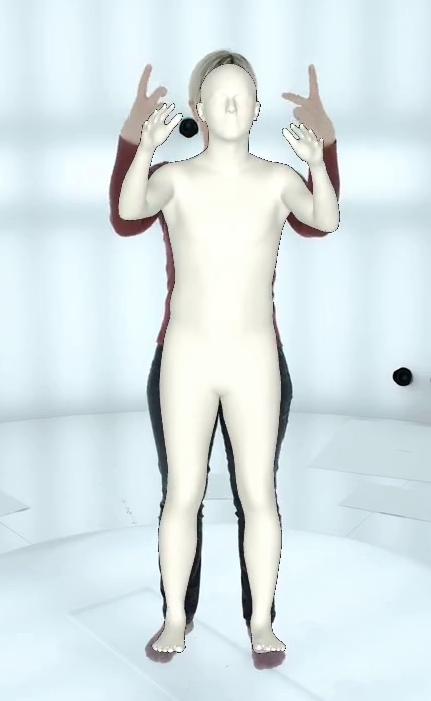}
                    \includegraphics[width=0.19\textwidth]{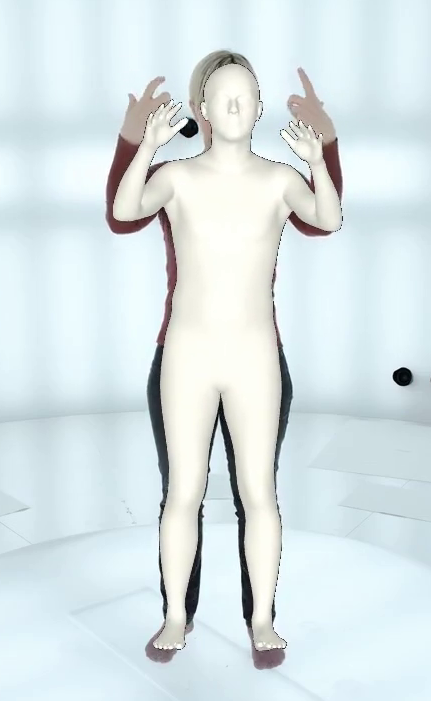}
                    \includegraphics[width=0.19\textwidth]{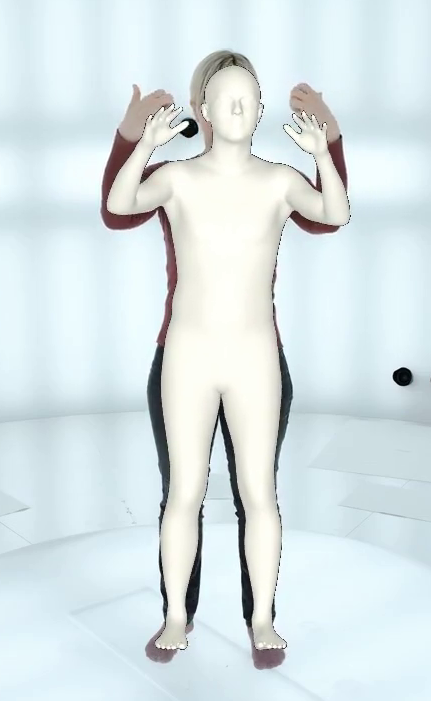}
                    \caption{VIBE results}
                \end{subfigure}
                
                \caption[Shape and Pose estimation for complete image sequence]{Shape and pose estimation results for an image sequence of own-data with sign language. SMPLify-X captures inconsistent body shape, severely disfigured arm poses and wobbling legs, while our work estimates the shape and motion stable and close to reality. VIBE is temporally stable but is not able to capture the complex hand motion.}
                \label{fig:no_wobble}
            \end{figure}

        Fig.~\ref{fig:no_wobble}, compares our approach to SMPLify-X as well as VIBE for a sequence which captures complex hand motion during sign language. While our approach achieves a consistent body shape, shape and pose estimated by SMPLify-X changes significantly, resulting in an inaccurate body shape. For the sake of completeness, we also show VIBE results here, but note that the comparison lacks as VIBE dose not include hands as well as expressive face. Hence, as expected, VIBE is not able to capture the complex hand motions depicted in the example frames.
        
\begin{figure}[t]
                \centering
                \includegraphics[width=0.2\linewidth]{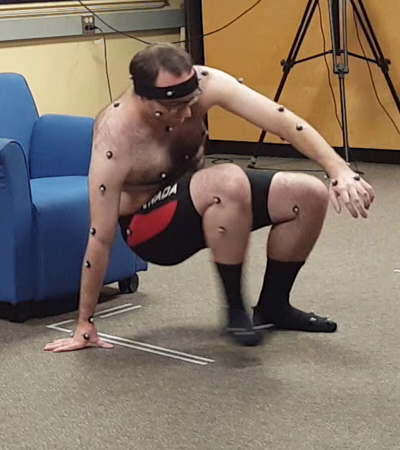}
                \includegraphics[width=0.75\linewidth]{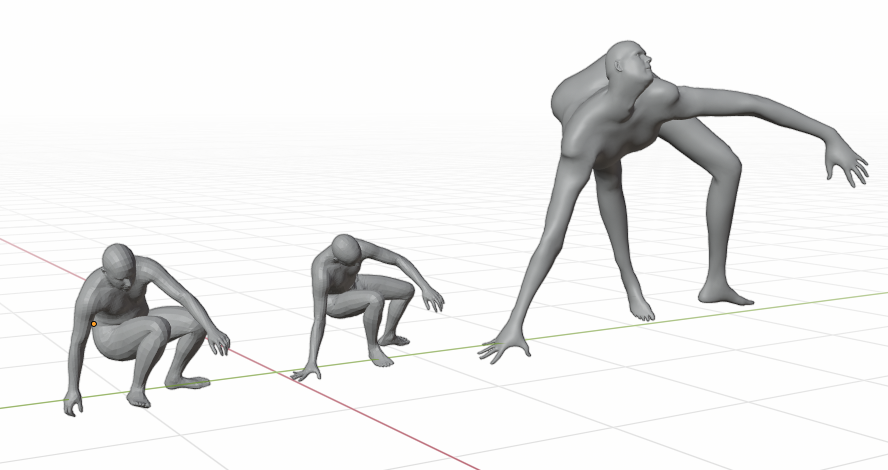}
                \caption[Shape estimation compared to ground truth]{The body model estimation meshes for the original frame on the left are shown on the right. The left body model shows the ground truth model, the central model shows the nearly accurate model estimate by our work's approach, while SMPLify-X's estimate can be seen on the right, having a completely distorted body shape.}
                \label{fig:comp_heights}
            \end{figure}

        Similarly for both Simple and Complex Motion, as they were defined in Sec.~\ref{sec:eval_setting}, our work derives more accurate estimates than SMPLify-X. This becomes apparent e.g. in Fig.~\ref{fig:comp_heights}, which shows subject 1 in the MoVi dataset performing Complex Motion, the ground truth body model, which was captured with AMASS and the body model estimates from our work and from SMPLify-X. It can be seen that SMPLify-X estimates a body shape and camera center resulting in a distorted body model being over 2m high in chrouching pose and more than 2.5 times as high as the ground truth model, while our work's method captures the body shape and pose accurately. 
        Moreover, in Fig.~\ref{fig:comp_heights}, one can see that the distance to the camera also remains consistent in time in our approach. This can be attributed to the forced shape consistency together with the keypoint reprojection.
        
        Empirically, we found that for random sampling, a sample size $|\mathcal{I}_{\textit{sub}}|$ of 15 is sufficient to find an accurately fitted shape, without being affected too much by inaccurate keypoints. 
        For increasingly complex motion with respect to occlusion, speed of motion, and so on, a higher sample size is recommended and for faster optimization a lower one.
An alternative sampling strategy is to choose the $|\mathcal{I}_{\textit{sub}}|$ images with the highest average keypoint confidence score, which improves results compared to random sampling with the same sample size. The improvement slope with higher sampling size of this strategy also decreases, because the newly added images have lower confidence scores than the ones already in use, and are thus less helpful in fitting the body shape.
            
            
            Our approach estimates one consistent and accurate shape for the whole image sequence, while SMPLify-X finds unrealistic shape changes within one motion sequence.

        \subsubsection{Volume Analysis}\label{sec:volume}
        The estimated shape is evaluated quantitatively by volume to volume comparison and with the mean volume error for an image sequence between the resulting body models of the method and the ground truth. The mean volume error for an image sequence is defined as the average of the per frame absolute difference between the volume of the estimated body model and the corresponding ground truth AMASS model.
        
\begin{figure}[ht]
            \centering
            \includegraphics[width=\linewidth]{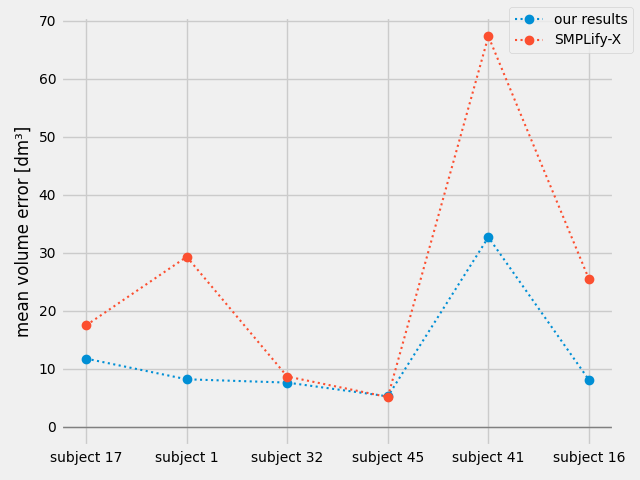}
            \caption[Mean volume error for Simple Motion]{Mean volume error for Simple Motion, subjects from MoVi.}
            \label{fig:volume_simple}
        \end{figure}

\begin{figure}[ht]
            \centering
            \includegraphics[width=\linewidth]{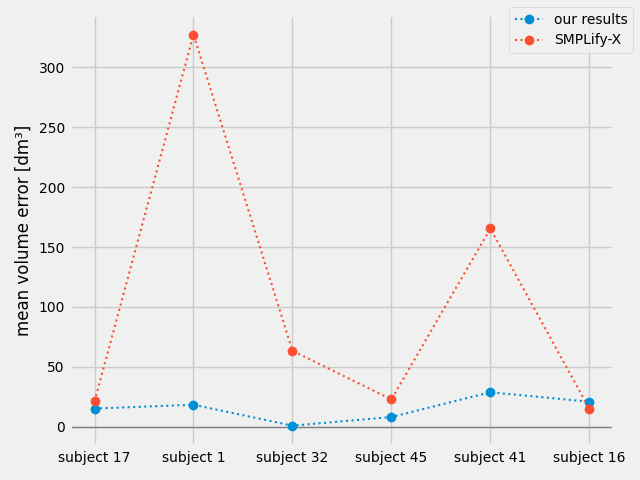}
            \caption[Mean volume error for Complex Motion]{Mean volume error for Complex Motion, subjects from MoVi.}
            \label{fig:volume_complic}
        \end{figure}

        The measured mean volume errors for our method and SMPLify-X are shown in Fig.~\ref{fig:volume_simple} for the Simple Motion sequences of the MoVi dataset. It can be seen that our method achieves a lower mean volume error than SMPLify-X for Simple Motion for all subjects. For Complex Motion, which is presented in Fig.~\ref{fig:volume_complic}, our work's approach produces consistent low mean volume errors, while SMPLify-X has extreme outliers, thus leading to severe deviations in estimated body shape. The data shows that our work achieves drastically better shape estimates than SMPLify-X.
        
        
\begin{figure}[htbp]
          \centering
              \subcaptionbox{Volumes of subject 1}{\includegraphics[width=0.95\linewidth]{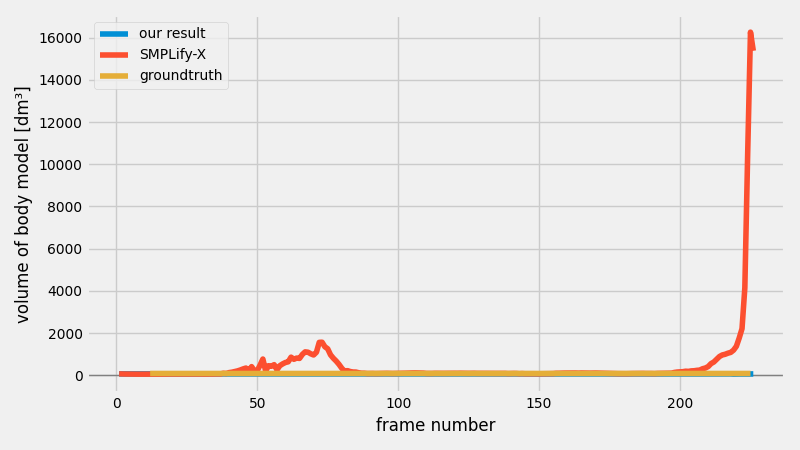}}
              \subcaptionbox{Volumes of subject 17}{\includegraphics[width=0.95\linewidth]{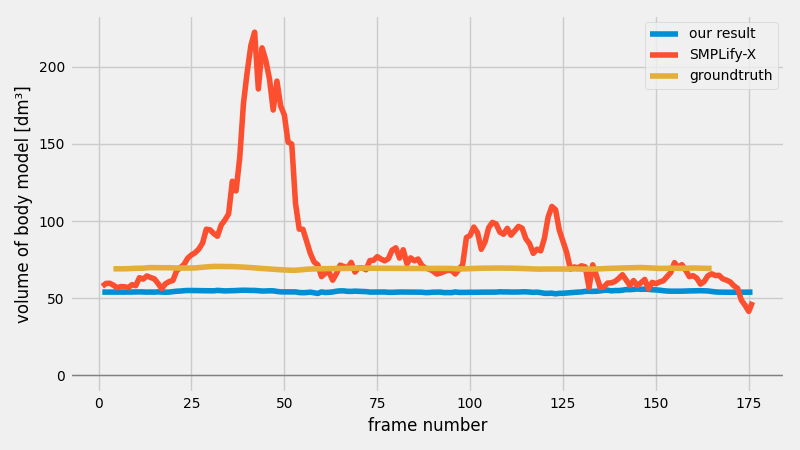}}
              \subcaptionbox{Volumes of subject 32}{\includegraphics[width=0.95\linewidth]{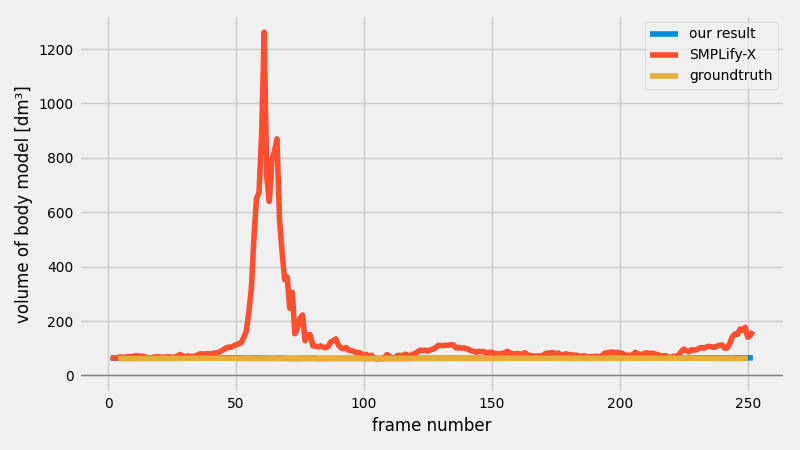}}
              \subcaptionbox{Volumes of subject 41}{\includegraphics[width=0.95\linewidth]{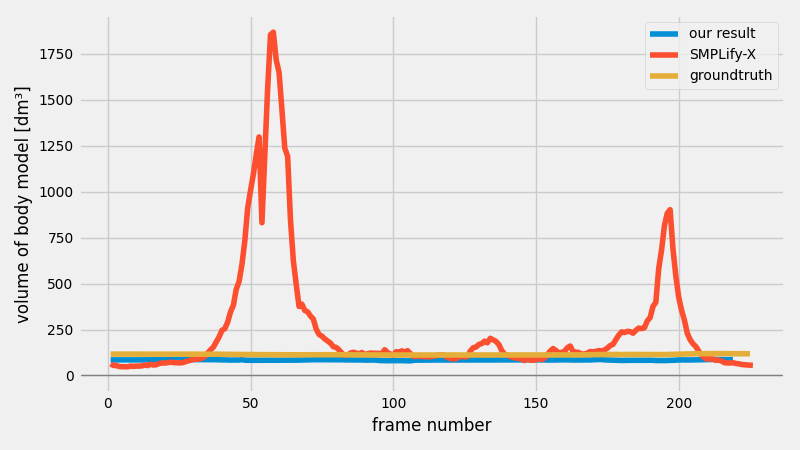}}
          \caption[Body volumes for Complex Motion]{Chronological sequences of body volume in body model estimation of Complex Motion. SMPLify-X results (red) have drastic deviations while our work's results (blue) are close to the ground truth (yellow).}
          \label{fig:volumes_smplifyx_peak}
        \end{figure}

        In Fig.~\ref{fig:volumes_smplifyx_peak}, the volume distribution over all frames of four subjects performing Complex Motion compared between our work, SMPLify-X, and the ground truth, is depicted.
       SMPLify-X produces body models with big deviations from ground truth and with the most inaccurate volumes in the first third and towards the end of the image sequence. One frame corresponding to the first peak for Complex Motion of subject 45 is shown in Fig.~\ref{fig:touch_ground_Subj_45} and one frame corresponding to the second peak in Fig.~\ref{fig:comp_heights}.
        Analysis of these frames with high SMPLify-X mean volume error
        reveals that this is the moment where the subjects put one hand on the floor to sit down. The other peak in the volume results for Complex Motion occurs at the moment where, while standing up, the last hand loses contact to the ground.
        Fig.~\ref{fig:volumes_smplifyx_peak} shows that our approach is in turn able to accurately and stably fit the body shape throughout the sequence, unaffected by these complex body poses.
        The reasoning behind some figures not having data for a few frames at the beginning or end of an image sequence, was given in Sec.~\ref{sec:eval_setting}.

        Because our approach calculates one shape for a whole image sequence and only allows changes of the pose for the individual frames, it is guaranteed that the shape is consistent over time and still adapted to the subject. In contrast to this, SMPLify-X calculates an individual body shape for each frame, which is independent of predecessor and successor frames and thus varies drastically from frame to frame in some instances, which became evident in this section's results.
        

        

\subsection{Temporal Consistency Evaluation}
        
        
        To evaluate the temporal consistency performance of our work, compared to SMPLify-X, quantitative and qualitative measures are applied in the following.
        
\subsubsection{Qualitative Analysis}
        
\begin{figure}[ht]
                \centering
                
                \vspace*{15mm}
                \begin{subfigure}[t]{0.325\linewidth}
                    \centering
                    \includegraphics[width=\linewidth]{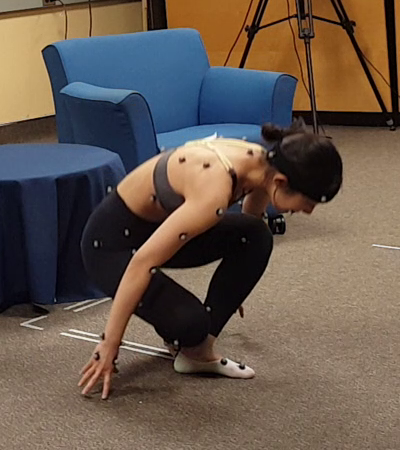}
                    \caption{Original frame}
                \end{subfigure}
                \begin{subfigure}[t]{0.325\linewidth}
                    \centering
                    \includegraphics[width=\linewidth]{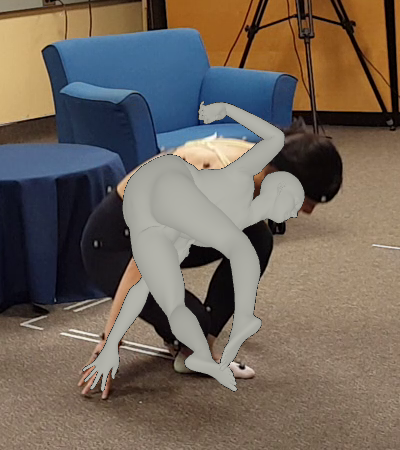}
                    \caption{SMPLify-X}
                \end{subfigure}
                \begin{subfigure}[t]{0.325\linewidth}
                    \centering
                    \includegraphics[width=\linewidth]{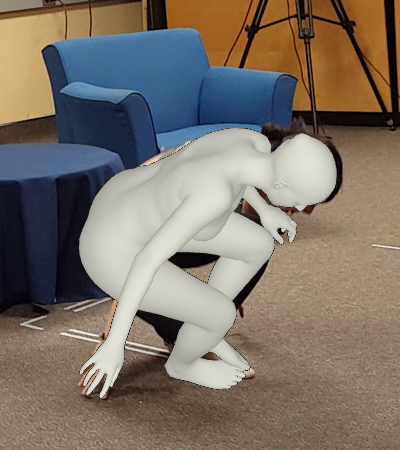}
                    \caption{Our approach}
                \end{subfigure}
                
                \caption[Pose estimation for Complex Motion]{Subject 45 from MoVi performing Complex Motion. SMPLify-X (b) is not able to capture the correct pose, but our work (c) estimates a nearly accurate body pose.}
            \label{fig:touch_ground_Subj_45}
            \end{figure}

        The introduction of temporal consistency improves results for image sequences containing subjects performing common and uncommon poses.
        Fig.~\ref{fig:touch_ground_Subj_45} shows a frame of the Complex Motion image sequence where a woman sits down on the floor (left) and the estimated body model from SMPLify-X (center) and our work's approach (right). Introducing temporal consistency the derived body model by our work captures the unusual body pose much better than SMPLify-X, which produces a body model with a distorted body pose.
        The body model in Fig.~\ref{fig:touch_ground_Subj_45}b is displayed darker than the one by our work, because SMPLify-X positioned the camera center and artificial light source inaccurately far away from the body model.
            

            \begin{figure}[ht]
                \centering
                
                \vspace*{15mm}
                \begin{subfigure}[t]{\linewidth}
                    \centering
                    \includegraphics[width=0.32\linewidth]{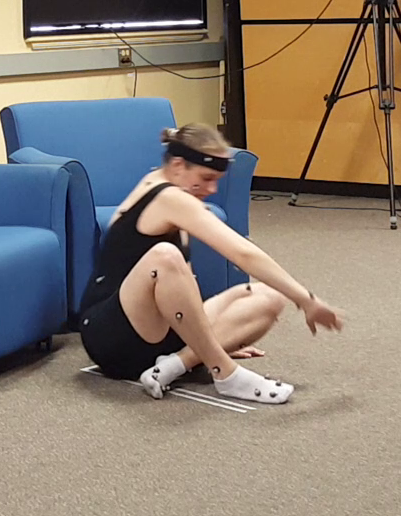}
                    \includegraphics[width=0.32\linewidth]{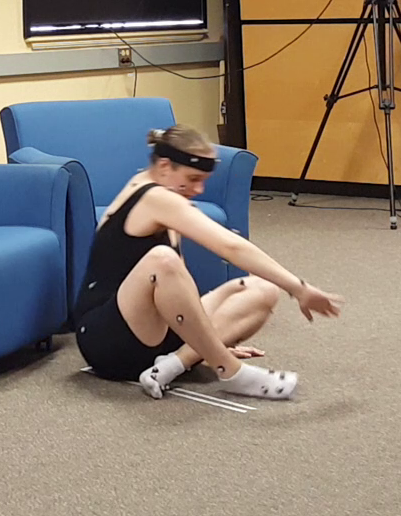}
                    \includegraphics[width=0.32\linewidth]{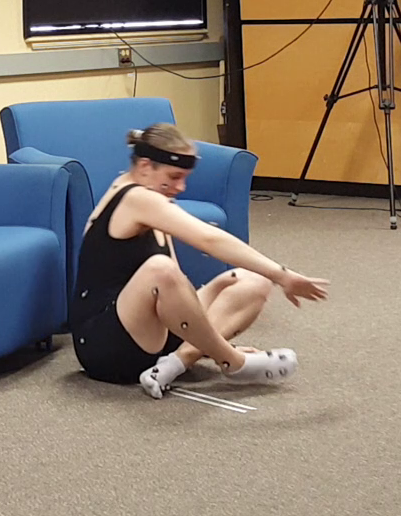}
                    \caption{Original frames}
                \end{subfigure}
                \vspace*{1mm}
                \begin{subfigure}[t]{\linewidth}
                    \centering
                    \includegraphics[width=0.32\linewidth]{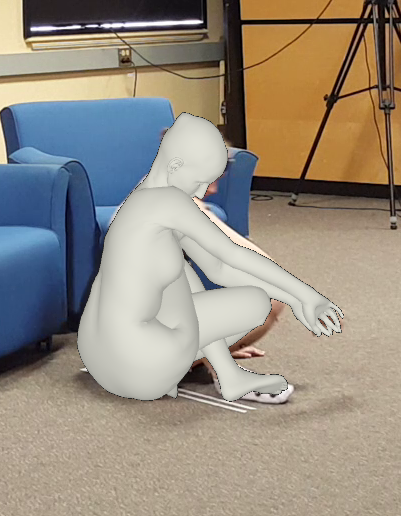}
                    \includegraphics[width=0.32\linewidth]{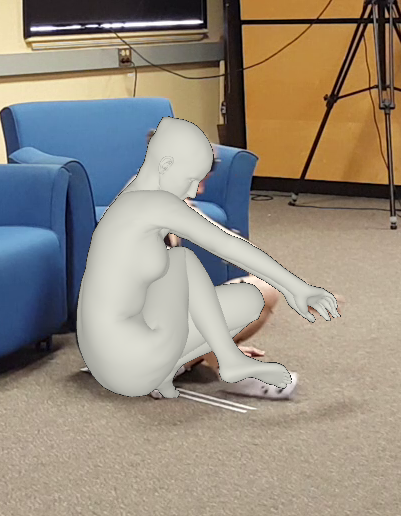}
                    \includegraphics[width=0.32\linewidth]{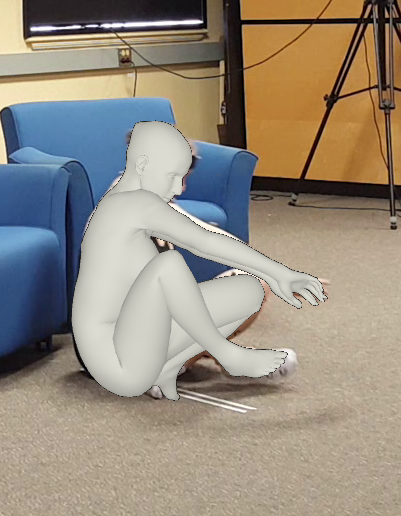}
                    \caption{Pose estimation results for window size 3}
                \end{subfigure}
                \vspace*{1mm}
                \begin{subfigure}[t]{\linewidth}
                    \centering
                    \includegraphics[width=0.32\linewidth]{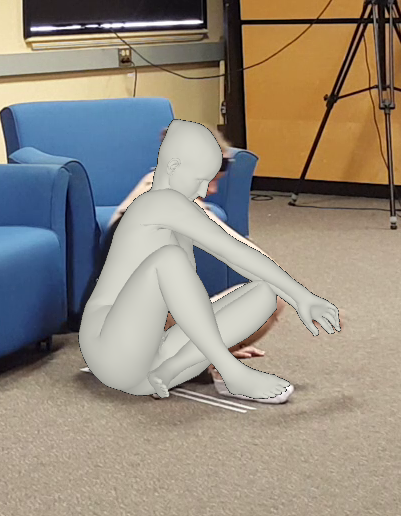}
                    \includegraphics[width=0.32\linewidth]{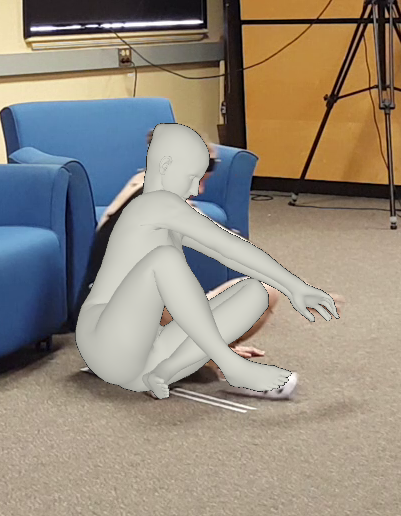}
                    \includegraphics[width=0.32\linewidth]{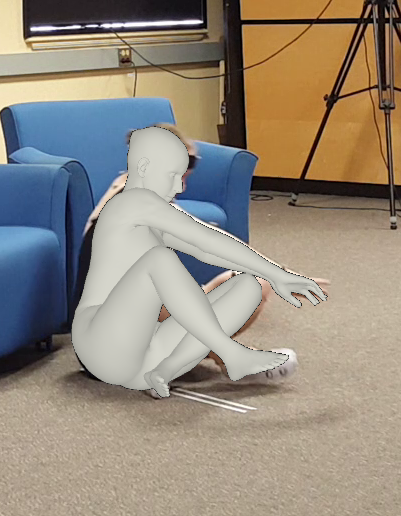}
                    \caption{Pose estimation results for window size 7}
                \end{subfigure}
                \caption[Pose estimation results for Complex Motion] {Subject 16 from MoVi performing Complex Motion (standing up from a cross-legged position) for window size 3 and 7. While a small window size of 3 does not sufficiently approximate the objective function in \ref{eq:overall_temp} and leads to instable results, a slightly larger window size of 7 leads to temporally stable and robust results.}
            \label{fig:touch_ground_Subj_16_ws}
            \end{figure}
      
        The video V1 in the Electronic Supplementary Material, shows the whole own-data sequence as well as results produced with our approach compared to SMPLify-X as well as VIBE. The first half of the video shows general hand movements and the second half shows sign language motion. Overall, our approach is able to prevent the jittering of motion and shape changes generated by SMPLify-X. 
        Furthermore, at the beginning, the hand movement estimates look similarly accurate; detailed results can be seen in Fig.~\ref{fig:no_wobble}. The results show that our approach prohibits extremely changing arm poses found in SMPLify-X, which do not describe the depicted motion, up to having both arms behind the head as well as wobbling of the legs noticeable for SMPLify-X. For complex hand motions of sign language, the impact of implementing temporal and shape consistency becomes very clear, since our approach is able to produce consistent results, but SMPLify-X fails to consistently produce accurate results. 
        
        The poses of hands and fingers are estimated rather accurately by both SMPLify-X and the presented approach, which also becomes apparent in Fig.~\ref{fig:no_wobble} and the referred video. In some cases the image based estimation captures hand and finger poses more accurately than the marker-based motion capture ground truth. Fig.~\ref{fig:comp_heights} shows this for the right hand of the subject. This is due to the subjects having had several markers on the hands, but not on the fingers. In contrast to SMPLify-X and our aproach, VIBE is not able to capture the complex hand motion.
            
            Performance evaluation of facial expression and jaw pose estimation poses problems, because the face usually only takes up a small part of the image and without high resolution it holds insufficient information for accurate keypoint and following body model estimation.
            Nonetheless, qualitative performance evaluation of jaw pose estimation revealed that neither SMPLify-X nor our work is able to reliably capture jaw pose correctly. In Fig.~\ref{fig:no_wobble} the closed jaw and facial expression was captured correctly, but often subjects with closed jaw are estimated by both approaches with an slightly open mouth, as can be seen intermittently in the second half of the referred video.
            Thus, both approaches capture hand and finger pose accurately, but are not able to reliably capture the jaw pose as well.
            
            The choice of the window size is a trade-off: An increasing window size leads to more accurate results, but also to a higher runtime, so it has to be chosen balancing those two goals. Nevertheless, as motion is not strongly correlated after a larger number of frames, experiments showed that even small window sizes of 7 or 9 are sufficient in order to produce stable and robust results. Fig.~\ref{fig:touch_ground_Subj_16_ws} shows example frames of Subject 16 from MoVi performing the complex motion of standing up from a cross-legged position. While a window size of 3 is not sufficient in order to approximate the objective function, a window size of 7 yields stable and robust results. Larger window sizes, however did not lead to a significant increase of quality in our experiments.
            
        Besides image and window size, the runtime depends on the complexity of the body motion. Hence, reported runtimes are only example numbers for orientation. On a PC as described in Sec.~\ref{sec:eval_setting}, the mean runtime was 0.154h per completed window, for images with size 720 x 960 pixels and window size of 7. For larger images of size 1920 x 1080, and window size 7, the mean runtime per window was 0.232h. Furthermore, runtime increases with more complex depicted body motion.

            
    Imposition of temporal consistency could be seen to enable improvements compared to SMPLify-X for usual and unusual body poses and to reduce the wobbling of estimated shape and motion.

        \subsubsection{Vertex-to-surface Error}
        
        For evaluation of the accuracy of the estimated body motion by SMPLify-X and our work's approach, the vertex-to-surface error to the ground truth meshes is compared. This error metric is computed for one body model to another with the iterative closest point algorithm, which translates and rotates the first mesh in such a way, that the distance between the meshes is minimized. This distance is calculated AABB-Tree based for every vertex of the first mesh to the closest point on the faces of the second mesh \cite{vertex-to-surface}. The average of all these distances is used as a measure of similarity between these body models.
        
\begin{figure}[ht]
            \centering
                \subcaptionbox{The mean vertex-to-surface error for Simple Motion}{
                    \includegraphics[width=\linewidth]{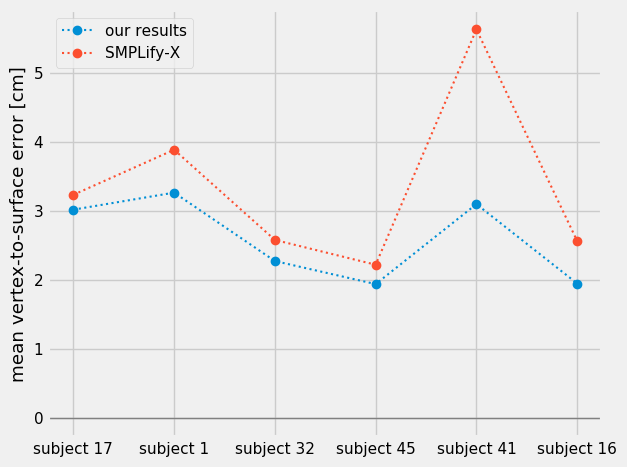}
                }
                \subcaptionbox{The mean vertex-to-surface error for Complex Motion}{
                    \includegraphics[width=\linewidth]{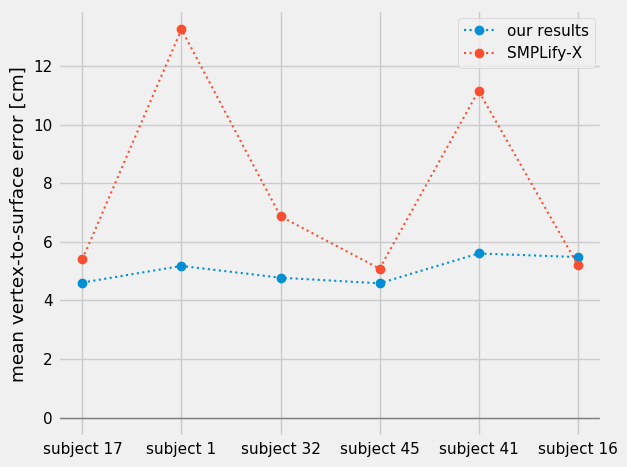}
                }
            \caption[Mean vertex-to-surface error for Simple and Complex Motion]{The mean vertex-to-surface error for all six subjects and both analyzed motions. Overall, our work's method achieves a lower mean vertex-to-surface error than SMPLify-X.}
            \label{fig:mean_v2s_plots}
        \end{figure}

        The vertex-to-surface error results for our work and SMPLify-X to the ground truth meshes can be seen in Fig.~\ref{fig:mean_v2s_plots}a for Simple and in Fig.~\ref{fig:mean_v2s_plots}b for Complex Motion.
        In 11 out of the 12 sequences our work's approach produces a lower vertex-to-surface error than SMPLify-X. 
        

        For further analysis of the vertex-to-surface error results, the mean and standard deviation for both approaches are given in Table~\ref{tab:vertex2surface_mean_and_stdev}.
        Mean and standard deviation of the vertex-to-surface error for both motions are significantly lower for our work's method than for SMPLify-X. 
        The estimation of the Simple Motion sequences undergoes improvement, but those of the Complex Motion sequences are especially strongly refined. 
        This shows a strong and comprehensive improvement in pose estimation by introducing shape and temporal consistency.
        
        \begin{table}[htb]
          \centering
           \begin{tabular}{c|cc}
           \textbf{Simple Motion} & mean & standard deviation \\
              \hline
              \hline
              \hline
              our work           & \textbf{2.59} cm & \textbf{0.81} cm             \\ 
              SMPLify-X       & 3.36 cm & 0.91 cm              
        \vspace{.5cm}\\
           \textbf{Complex Motion} & mean & standard deviation \\
              \hline
              \hline
              \hline
              our work           & \textbf{5.04} cm & \textbf{2.52} cm            \\ 
              SMPLify-X       & 7.83 cm & 6.62 cm             \\ 
            \end{tabular}
        \caption[Mean and standard deviation of vertex-to-surface error]{Mean and standard deviation of the vertex-to-surface error for both methods and motions.}
        \label{tab:vertex2surface_mean_and_stdev}
        \end{table}

The results of the evaluation section are discussed in the following.
        

\section{Discussion}\label{sec:discussion}


In this work, temporal and shape consistency was imposed on body shape and motion estimation from monocular image sequences. This was achieved in a three-phased optimization framework, based on the approach presented in SMPLify-X \cite{SMPLify-X}.
Quantitative and qualitative analysis of body model estimation revealed that our approach improves results in comparison to SMPLify-X.
By imposing temporal and shape consistency, the body shape is derived more accurately and the body pose estimation is improved significantly.

We showed that the body shape of a depicted subject can be estimated accurately with the optimization in the shape fitting phase (Sec.~\ref{sec:shapefitting}). Quantitative evaluation results showed that our approach brings more accurate shape estimates than SMPLify-X, not only for simple, but also for complex body motion and poses. However, a deviation of over $30dm^3$ was detected for subject 41 (see Fig.~\ref{fig:volume_simple}), who was dressed completely in black.
To address this, further work could research deriving the body shape from the body silhouette, which could be found using image segmentation.

Evaluation of the imposition of temporal consistency on body motion, in addition to shape consistency, revealed that our approach enables accurate derivation of complex body poses and partially self-occluded bodies.
It was able to resolve the wobbling of shape and motion which was seen for SMPLify-X's results.
Instead, our approach resulted in smooth body motion, which was also quantitatively shown to be closer to the ground truth meshes.

In addition, we discovered that imposing shape and temporal consistency together leads to a temporally consistent behaviour of the distance between body model and camera. 
The resulting effect is a consistent, realistic distance between the model and the camera for our approach, whereas SMPLify-X generates inconsistent distances to the camera, including noticeable outliers.

\section{Conclusion}\label{sec:conclusion}

We applied our approach for accurate estimation of body shape and motion to challenging tasks, such as estimation of intricate 
body motion or sign language capture, where correct timing and smooth motion are extremely important for both expressiveness as well as acceptance.
Based on quantitative and qualitative analysis, it can be concluded that imposing shape and temporal consistency significantly improves the accuracy of estimated body models. 
Therefore, it is e.g. suitable for the task of generating training data for realistic sign language reproduction from real image sequences or videos.
This enables sign language support for arbitrary videos using artificial avatars, allowing a more inclusive access to many forms of media.
Accurate body shape and motion estimation can furthermore be used in a variety of other contexts, like understanding of human social behavior or body motion tracking for the entertainment industry, robotics, sports analytics, medical applications and many more.

As could be seen in some qualitative and quantitative evaluation, SMPLify-X is able to derive an accurate body shape and pose from images in many cases but shows inaccuracies for some scenarios and a general jittering of body motion, since it processes every frame without the rest of the image sequence as context. These inaccuracies and jittering were dissolved to a great extent by the approach presented in this paper.
The introduction of temporal and shape consistency to body pose estimation thus enables capturing motion realistically steady and smooth.


\section{Acknowledgements}{This work has partly been funded by the European Union’s Horizon 2020 research and innovation programme under agreement No 952147 (Invictus) as well as the German Federal Ministry of Education and Research (BMBF) through the Research Program MoDL under Contract no. 01 IS 20044.

The employed data set MoVi is available under \cite{MoVi}, the data set own-data and the code produced for this paper are not publicly available. There are no competing interests and no further acknowledgements regarding this publication. All authors have contributed sufficiently to the scientific work.
}

\bibliographystyle{CVM}


\end{document}